\pdfoutput=1
\documentclass[11pt]{article}

\usepackage[preprint]{acl}

\usepackage{times}
\usepackage{latexsym}

\usepackage[T1]{fontenc}
\usepackage[utf8]{inputenc}

\usepackage{microtype}

\usepackage{inconsolata}

\usepackage{graphicx}

\usepackage[most]{tcolorbox} % coloured boxes
\usepackage[table,dvipsnames]{xcolor} % special colours
\usepackage{hyperref} % to use href (needed!)
\usepackage{siunitx} % to round numbers automatically
\newcommand{\pnum}[1]{\begingroup\sisetup{round-precision=3}\num{#1}\endgroup} % this is to round p values (3 decimal places)
\usepackage{booktabs,tabularx,threeparttable,enumitem}

\title{If Probable, Then Acceptable? Understanding Conditional Acceptability Judgments in Large Language Models}

\author{
Jasmin Orth\hspace{0.1mm}\thanks{Equal contribution.}\textsuperscript{\normalfont 1}\hspace{6mm}Philipp Mondorf\footnotemark[1]\textsuperscript{\normalfont 1, 2}\hspace{6mm}Barbara Plank\textsuperscript{\normalfont 1, 2}\\
$^1$MaiNLP, Center for Information and Language Processing, LMU Munich, Germany\\
$^2$Munich Center for Machine Learning (MCML), Munich, Germany\\
{\tt\footnotesize jasmin.orth@campus.lmu.de, \{p.mondorf, b.plank\}@lmu.de} \\
}

\begin{document}
\maketitle
\begin{abstract}
Conditional acceptability refers to how plausible a conditional statement is perceived to be. It plays an important role in communication and reasoning, as it influences how individuals interpret implications, assess arguments, and make decisions based on hypothetical scenarios. When humans evaluate how acceptable a conditional ``If A, then B'' is, their judgments are influenced by two main factors: the \emph{conditional probability} of $B$ given $A$, and the \emph{semantic relevance} of the antecedent $A$ given the consequent $B$ (i.e., whether $A$ meaningfully supports $B$). While prior work has examined how large language models (LLMs) draw inferences about conditional statements, it remains unclear how these models judge the \emph{acceptability} of such statements. To address this gap, we present a comprehensive study of LLMs' conditional acceptability judgments across different model families, sizes, and prompting strategies. Using linear mixed-effects models and ANOVA tests, we find that models are sensitive to both conditional probability and semantic relevance\textemdash{}though to varying degrees depending on architecture and prompting style. A comparison with human data reveals that while LLMs incorporate probabilistic and semantic cues, they do so less consistently than humans. Notably, larger models do not necessarily align more closely with human judgments.
\end{abstract}

\section{Introduction}
Conditional statements of the form ``\emph{If $A$, then $B$}'' are fundamental to how humans make predictions, draw conclusions, and offer explanations~\citep{byrne2007imagination, evans2004conditionalsIF, douven2015epistemology}. As such, they play a central role in human communication, reasoning, and decision-making. 
Understanding the conditions under which humans judge conditional statements as \emph{acceptable} or \emph{plausible} sheds light on the principles that underlie everyday language use, from casual conversation to scientific explanations.
Consider the following example, drawn from~\citet{skovgaard2016relevance}:

\vspace{1mm}
\noindent (i) \emph{If Mark pulls the plug on his TV, then it will be turned off.}\\
\noindent (ii) \emph{If Mark is wearing socks, then his TV will be working.}
\vspace{1mm}

\begin{figure}[t]
  \includegraphics[width=\columnwidth]{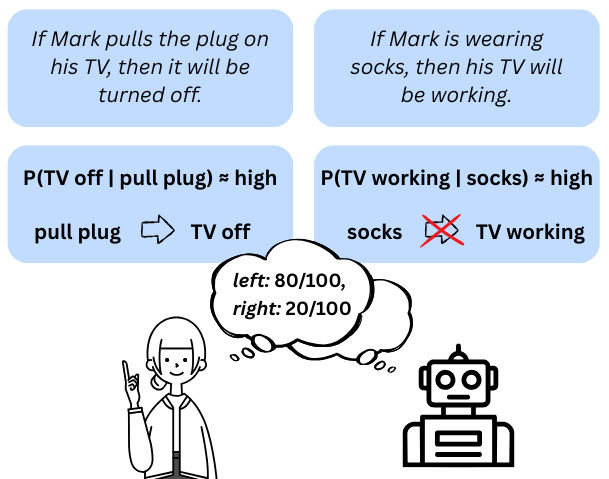}
  \caption{Illustration of two conditionals with equally high conditional probabilities but differing evidential relevance. While both are probable, only the left one encodes a plausible causal or evidential link between antecedent and consequent, appearing more acceptable.}
  \label{fig:intro}
\end{figure}

While both sentences are grammatically sound, the second appears less acceptable: it is unlikely that Mark wearing socks has any effect on whether his TV is working.
This notion of \emph{acceptability}, or how appropriate a conditional statement seems in a given context, is often judged intuitively by humans~\citep{douven2012indicatives, skovgaard2016relevance}. Two main hypotheses aim to explain such intuitions. The \emph{conditional probability hypothesis} holds that acceptability depends primarily on the \emph{conditional probability} of $B$ given $A$~\citep{evans2003conditionals, over2007probability}. By this account, both conditionals above might be judged acceptable, as the probability of the TV being turned off given that the plug has been pulled, and the probability of the TV working given that Mark is wearing socks, are both high.
In contrast, the \emph{evidentiality hypothesis} emphasizes an additional influence of \emph{relevance} or \emph{evidential connection} encoded within the statement~\citep{douven2012indicatives, skovgaard2016relevance}. From this perspective, (i) is acceptable because pulling the plug is causally connected to the TV turning off, whereas (ii) is not, because wearing socks lacks such a connection.
LLMs have demonstrated notable capabilities across diverse linguistic and cognitive tasks, including general reasoning~\citep{hao2024llm, mondorf2024beyond}, linguistic inference~\citep{clark2020transformers}, and causal reasoning~\citep{dettki2025large, yu2024improving}. 
However, despite increasing work on LLM reasoning, research on conditional \emph{acceptability} judgments in LLMs remains virtually nonexistent. 
Most existing studies on LLMs and conditional statements focus on their causal, abductive, or inferential capabilities~\citep{dettki2025large, holliday2024conditional, liu2023magic, mondorf-plank-2024-comparing}, rather than on whether conditionals are judged contextually \emph{acceptable}.

We address this gap by conducting a targeted investigation of conditional acceptability judgments in LLMs. Building on prior work from cognitive psychology~\citep{skovgaard2016relevance}, we assess the acceptability judgments of two model families: Llama 3.1 (8B and 70B) and Qwen 2.5 (7B and 72B), and examine how their judgments vary as a function of both the conditional probability and the evidential connection encoded within a statement.
Our findings reveal that:

\begin{itemize}
    \item Acceptability judgments of models are influenced by both conditional probability and semantic relation, though the strength and consistency of this effect vary substantially by model family, size, and prompting technique.
    \item Comparison with human data suggests that while the overall trends in LLM judgments broadly resemble those of humans, LLMs show greater variability across prompts and model sizes and a less \emph{systematic} integration of probabilistic and semantic cues.
    \item Larger model size does not necessarily lead to more alignment with human trends, and while few-shot prompting can enhance LLMs' sensitivity to semantic relevance, it may introduce bias toward specific semantic relations.
\end{itemize}

\section{Background}

\subsection{Conditional Acceptability}
The key concept of this study is \emph{conditional acceptability}: the degree to which a conditional statement is perceived to be appropriate, plausible, or natural in context~\citep{skovgaard2016relevance}. While often associated with \emph{conditional probability} $P(B \mid A)$, such as proposed by the \emph{conditional probability hypothesis}~\citep{evans2003conditionals, oaksford2003conditional, over2007probability}, the two can diverge significantly in practice.
Take, for instance, the statement from the introduction: \emph{If Mark is wearing socks ($A$), then his TV will be working ($B$).}
This may be judged unacceptable or odd, even when both $P(A)$ and $P(B)$ are high, and even when $P(B \mid A)$ is high. The issue lies not in probability, but in the perceived irrelevance of $A$ to $B$. The statement violates intuitive notions of coherence or causality: wearing socks does not cause the TV to function, nor would not wearing them affect the outcome. As such, the distinction reflects the broader gap between \emph{correlation} and \emph{causation}~\citep{barrowman2014correlation}.
This disconnect has prompted extensive psychological research on how humans evaluate conditional statements~\citep[e.g.,][]{douven2015epistemology, evans2004conditionalsIF, hardman2003thinking}, suggesting that individuals often rely on a combination of semantic, contextual, and probabilistic cues when assessing conditionals~\citep{berto2021indicative, douven2012indicatives, over2007probability}.
Specifically, the \emph{evidentiality hypothesis} argues that the semantic relation between antecedent $A$ and consequent $B$\textemdash{}whether $A$ supports, contradicts, or is irrelevant to $B$\textemdash{}plays a crucial role in human acceptability judgments~\citep{douven2012indicatives, skovgaard2016relevance, berto2021indicative}. Returning to the earlier example, \emph{If Mark is wearing socks, then his TV will be working} may be judged unacceptable not because $P(B \mid A)$ is low, but because there is no plausible explanatory or causal (\emph{supporting}) link between the two clauses.
In particular,~\citet{skovgaard2016relevance} find that humans generally judge conditionals with a \emph{supporting} relation more acceptable than those with a \emph{contradicting} or \emph{irrelevant} relation.

Although studies have investigated how \emph{humans} judge the acceptability of conditional statements, it remains unclear how and under what circumstances LLMs judge conditionals as acceptable.

\begin{figure}[tbp]
  \small
  \centering
  \begin{tcolorbox}[
    colback=PineGreen!5!white,
    colframe=PineGreen,
    boxrule=0.4pt,
    rounded corners,
    width=\columnwidth,
    fonttitle=\bfseries,
    before upper={\parindent0pt}
  ]
  \textbf{Positive relation ($POS$):}
  
  If Nicole sunbathes on the beach, then she will get sunburned.
  \end{tcolorbox}
  \begin{tcolorbox}[
    colback=PineGreen!5!white,
    colframe=PineGreen,
    boxrule=0.4pt,
    rounded corners,
    width=\columnwidth,
    fonttitle=\bfseries,
    before upper={\parindent0pt}
  ]
  \textbf{Negative relation ($NEG$):}
  
  If Nicole sunbathes on the beach, then her skin will continue to be pale.
  \end{tcolorbox}
  \begin{tcolorbox}[
    colback=PineGreen!5!white,
    colframe=PineGreen,
    boxrule=0.4pt,
    rounded corners,
    width=\columnwidth,
    fonttitle=\bfseries,
    before upper={\parindent0pt}
  ]
  \textbf{Irrelevant relation ($IRR$):}
  
  If Nicole knows what a computer is, then she will get sunburned.
  \end{tcolorbox}
  \caption{Example of three conditional statements representing different relations.}
  \label{fig_scenario_example_Nicole_all}
\end{figure}

\subsection{Dataset} \label{background_dataset}
To investigate how LLMs assess the acceptability of conditional statements, we draw on a dataset from ~\citet{skovgaard2016relevance}. Their study aims to understand how humans judge conditional acceptability using 144 conditional statements embedded in 12 short, realistic contexts.\footnote{More information about the dataset and demographics can be found in Appendix~\ref{app_human_study}.} Each context is tied to a unique everyday scenario, such as going on a date or attending a meeting.

To enable systematic analyses, the 12 conditionals that share the same context are organized along two dimensions. First, each conditional can be classified into one of three \emph{relation types}, indicating the semantic connection between antecedent $A$ and consequent $B$: a \textbf{positive} relation ($POS$) means $A$ supports $B$, a \textbf{negative} relation ($NEG$) denotes that $A$ undermines or contradicts $B$, and an \textbf{irrelevant} relation ($IRR$) means $A$ is unrelated to $B$ (see Figure~\ref{fig_scenario_example_Nicole_all}). Second, the conditionals vary in their \emph{prior probability}: each scenario includes four levels based on combinations of high or low probability for both antecedent $P(A)$ and consequent $P(B)$ (e.g., high-high, high-low, low-high, low-low; see Appendix~\ref{app_dataset} for details). This yields 12 unique configurations per context (3 relation types $\times$ 4 prior probabilities, as shown in Table~\ref{tab:config_matrix} in the Appendix), resulting in 144 conditionals in total.
The following example illustrates how relation types can be defined within a given context:

\vspace{1mm}
\emph{Nicole’s skin is pale and she wants a suntan with a mild brown color. It's summer, the weather is nice, and there is a beach nearby. However, Nicole’s skin is sensitive to the sun, and she is bad at controlling the amount of her exposure to the sun.}
\vspace{1mm}

As shown in Figure~\ref{fig_scenario_example_Nicole_all}, a \textbf{positive} relation is illustrated by the conditional statement: \emph{If Nicole sunbathes on the beach, then she will get sunburned}. Here, under the contextual constraint of her having sensitive skin, \emph{Nicole sunbathing} actively causes her to \emph{get sunburned}. In the \textbf{negative} example, the antecedent \emph{Nicole sunbathing on the beach}  contradicts her \emph{staying pale}, while the \textbf{irrelevant} example pairs two unrelated ideas: \emph{knowing what a computer is} has no causal connection to \emph{getting sunburned}.
These structured scenarios are designed to help participants apply real-world knowledge while simultaneously constraining interpretation, anchoring relations and probability judgments to a shared context. In the present study, we use this dataset to evaluate how LLMs judge the acceptability of conditionals and whether they exhibit similar patterns as humans. By prompting models with these same context-rich scenarios and conditionals, we can systematically assess their responses across relation types and probability levels.

\section{Method}

\subsection{Experimental Setup}\label{exp_setup}

\paragraph{Tasks}

We follow the approach of~\citet{skovgaard2016relevance} and present the 144 conditional statements described in Section~\ref{background_dataset} to LLMs in three distinct judgment tasks:
(i) estimating the conditional probability $P(B \mid A)$ of the consequent given the antecedent, (ii.a) predicting the overall probability of the conditional $P\left(\text{``\emph{If A, then B}''}\right)$, and (ii.b) judging the \emph{acceptability} of the conditional statement ``\emph{If A, then B}''. While tasks (ii.a) and (ii.b) present the LLMs with full conditional statements (e.g., ''\emph{If Mark is wearing socks, then his TV will be working}''), the conditional probability task (i) presents the statement in a split format, consisting of a supposition such as ``\emph{Suppose Mark is wearing socks}'' and an outcome, such as ''\emph{His TV will be working}'', which needs to be judged (see Figure~\ref{fig_statement_versions_mark} lower bounding box).

\begin{figure}[tbp]
  \small
  \centering
  \begin{tcolorbox}[
    colback=MidnightBlue!5!white,
    colframe=MidnightBlue,
    boxrule=0.4pt,
    rounded corners,
    width=\columnwidth,
    before upper={\parindent0pt}
  ]
  \textbf{Statement:}

  If Mark is wearing socks, then his TV will be working.
  \end{tcolorbox}
  \begin{tcolorbox}[
    colback=PineGreen!5!white,
    colframe=PineGreen,
    boxrule=0.4pt,
    rounded corners,
    width=\columnwidth,
    fonttitle=\bfseries,
    before upper={\parindent0pt}
  ]
  \textbf{Supposition:}

  Suppose Mark is wearing socks.
  
  \textbf{Outcome:}
  
  His TV will be working.
  \end{tcolorbox}
  \caption{Example of a conditional presented in full form (top) and its corresponding split version (bottom), used to elicit different types of model judgments.}
  \label{fig_statement_versions_mark}
\end{figure}

To analyze whether LLMs' acceptability judgments align with a probabilistic or relevance-based interpretation, these tasks are designed to separate probabilistic reasoning from pragmatic or communicative considerations. The conditional probability task described above provides a reference point reflecting pure probabilistic reasoning: if large language models rely solely on conditional probability when judging conditional statements, their \emph{if-probability}\footnote{To make the distinction between the conditional probability $P(B \mid A)$ and the probability of the full conditional statement as clear as possible, we denote the probability of the full conditional statement the \emph{if-probability} in this study.} and acceptability ratings would be expected to closely match their corresponding $P(B \mid A)$ values.
The split between acceptability and if-probability judgments further allows to determine whether assessments of conditional statements differ depending on the type of judgment.
All tasks instruct the model to respond with a single number between $0-100$, indicating the degree of probability or acceptability. % For a discussion on extracting probability judgments through direct elicitation as opposed to other techniques such as sentence probability, please refer to Appendix~\ref{app_elicitation_methods}. 
Full prompt templates are provided in Figures~\ref{fig_llm_prompt_vanilla_cprob} to~\ref{fig_llm_prompt_vanilla_ifacc} in Appendix~\ref{app_prompt_templates}.

\begin{figure*}[tbp] % page 5
  \centering
  
  \includegraphics[width=0.32\linewidth]{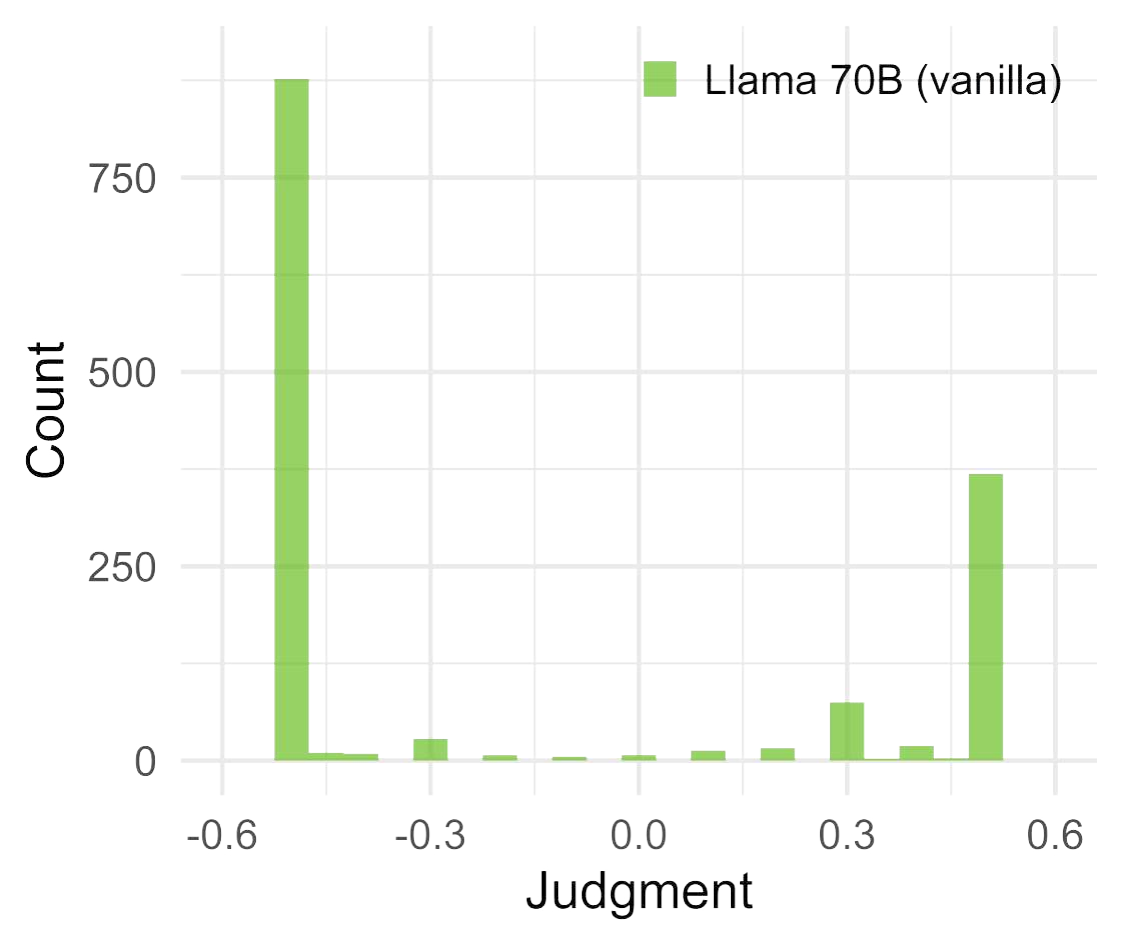}
  \includegraphics[width=0.32\linewidth]{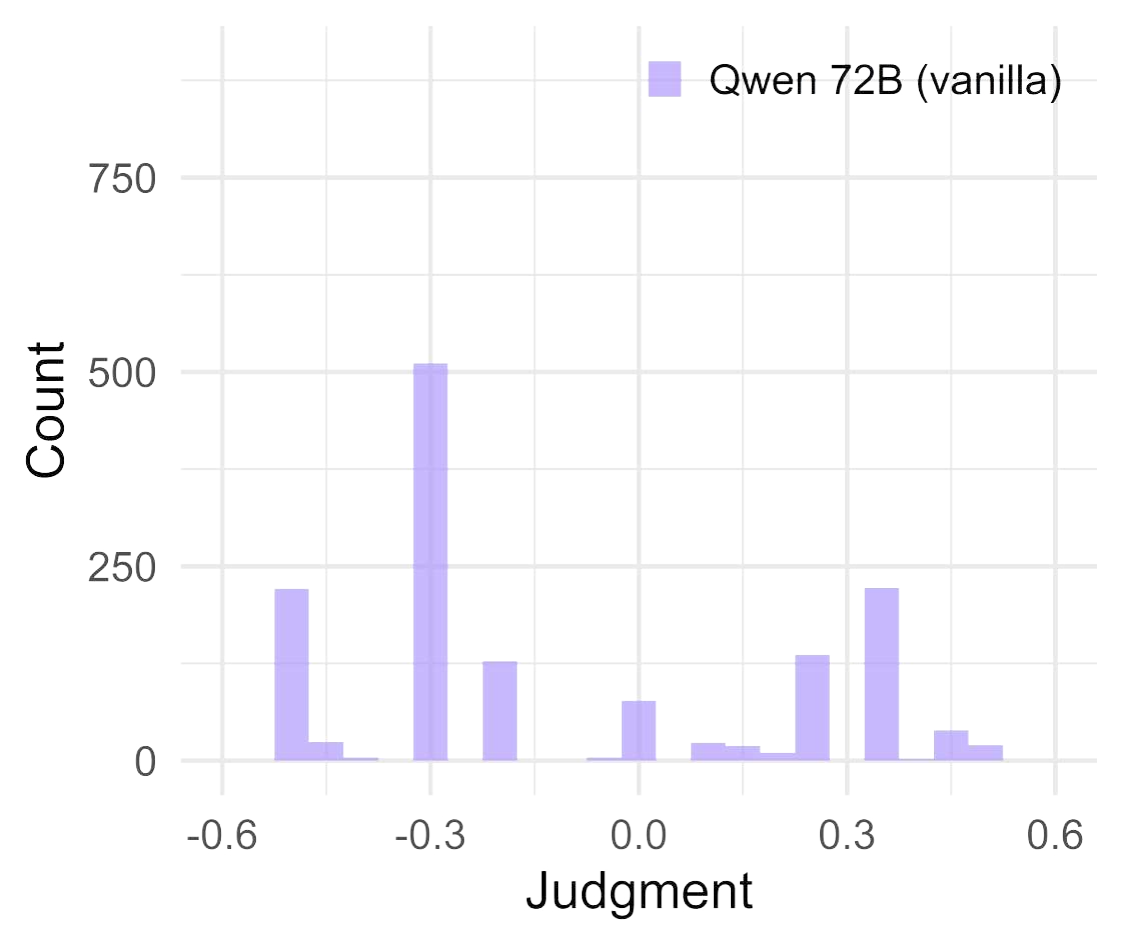}
  \includegraphics[width=0.32\linewidth]{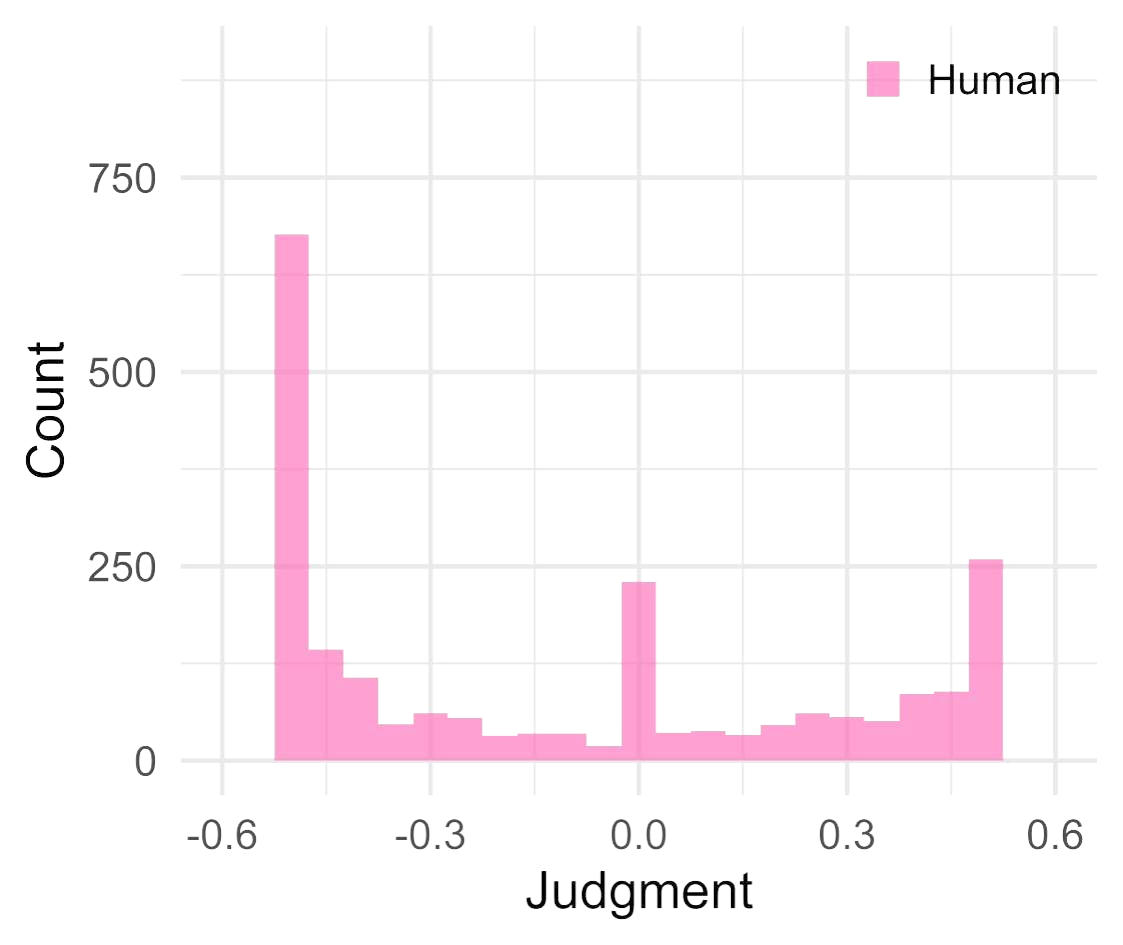}

  \vspace{1em}

  \includegraphics[width=0.32\linewidth]{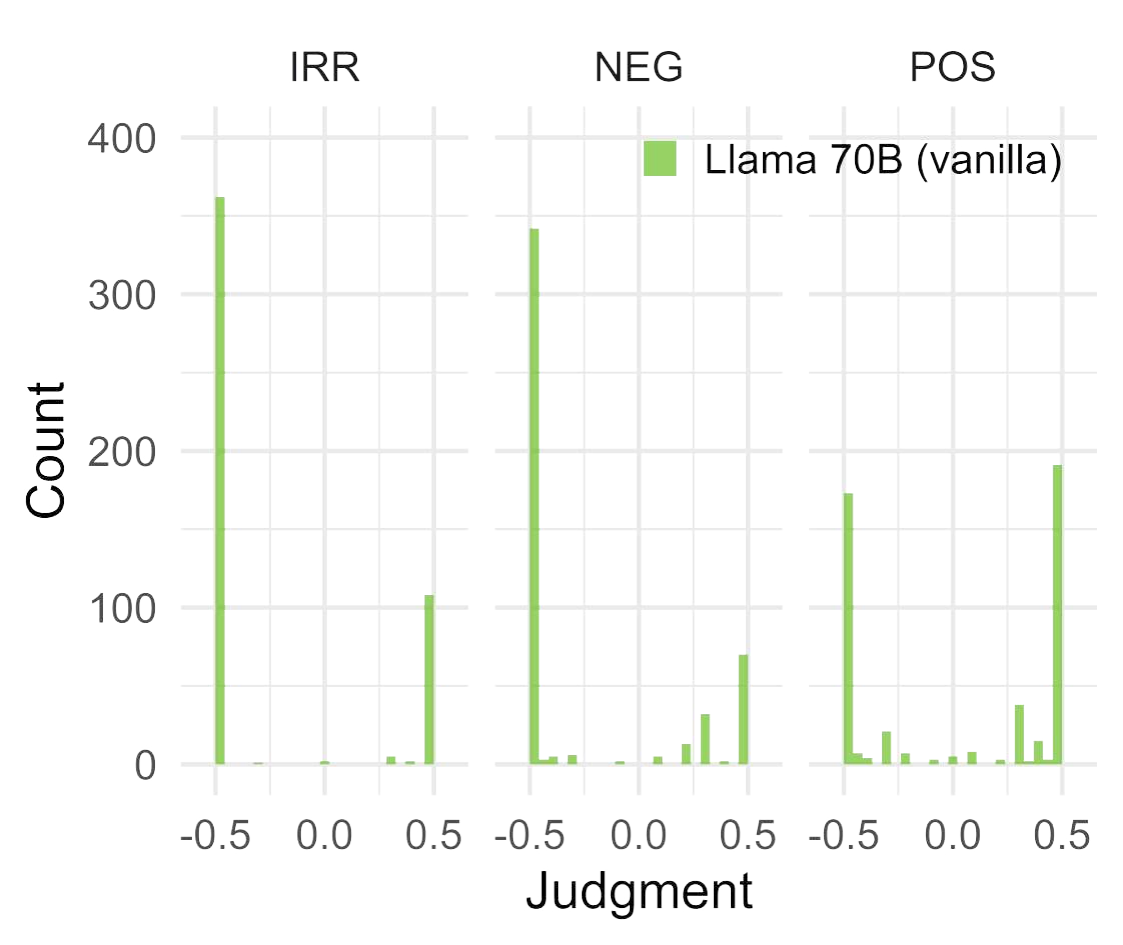}
  \includegraphics[width=0.32\linewidth]{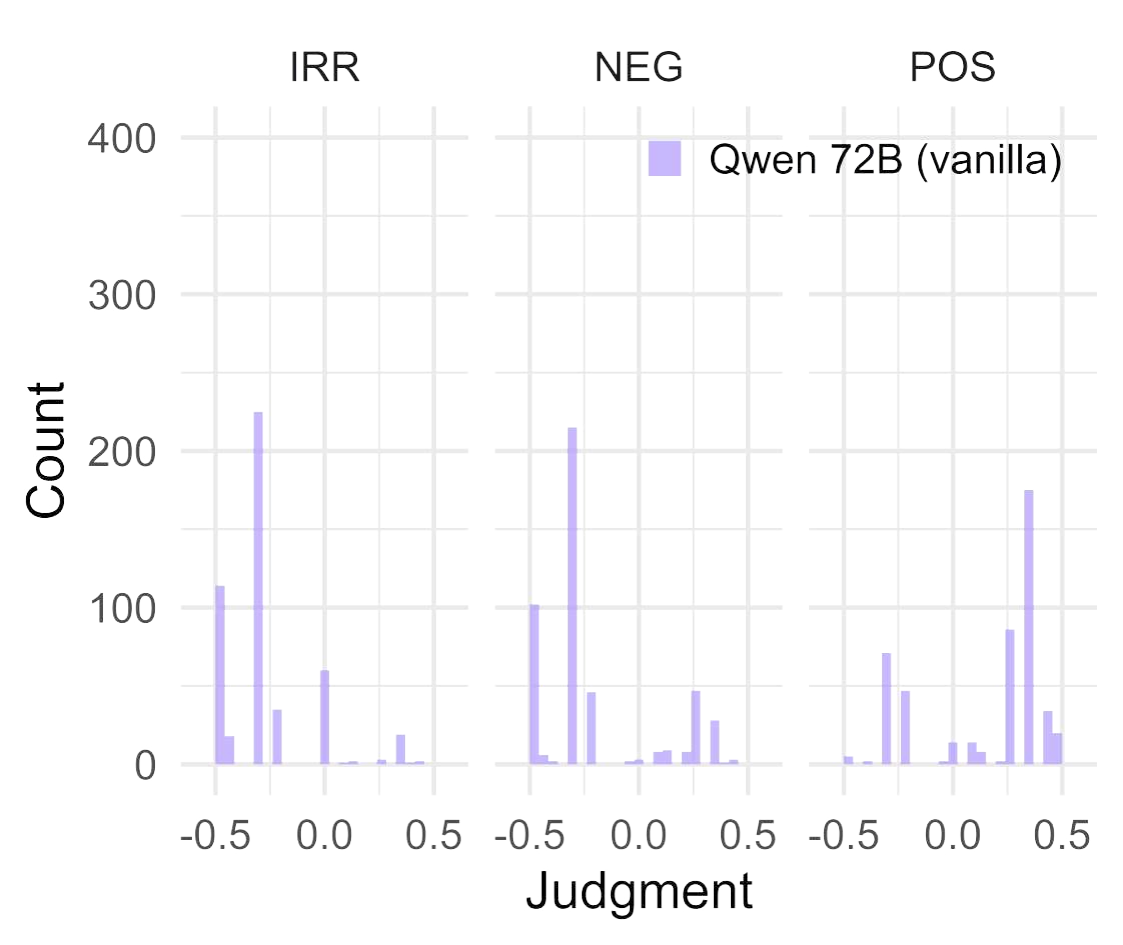}
  \includegraphics[width=0.32\linewidth]{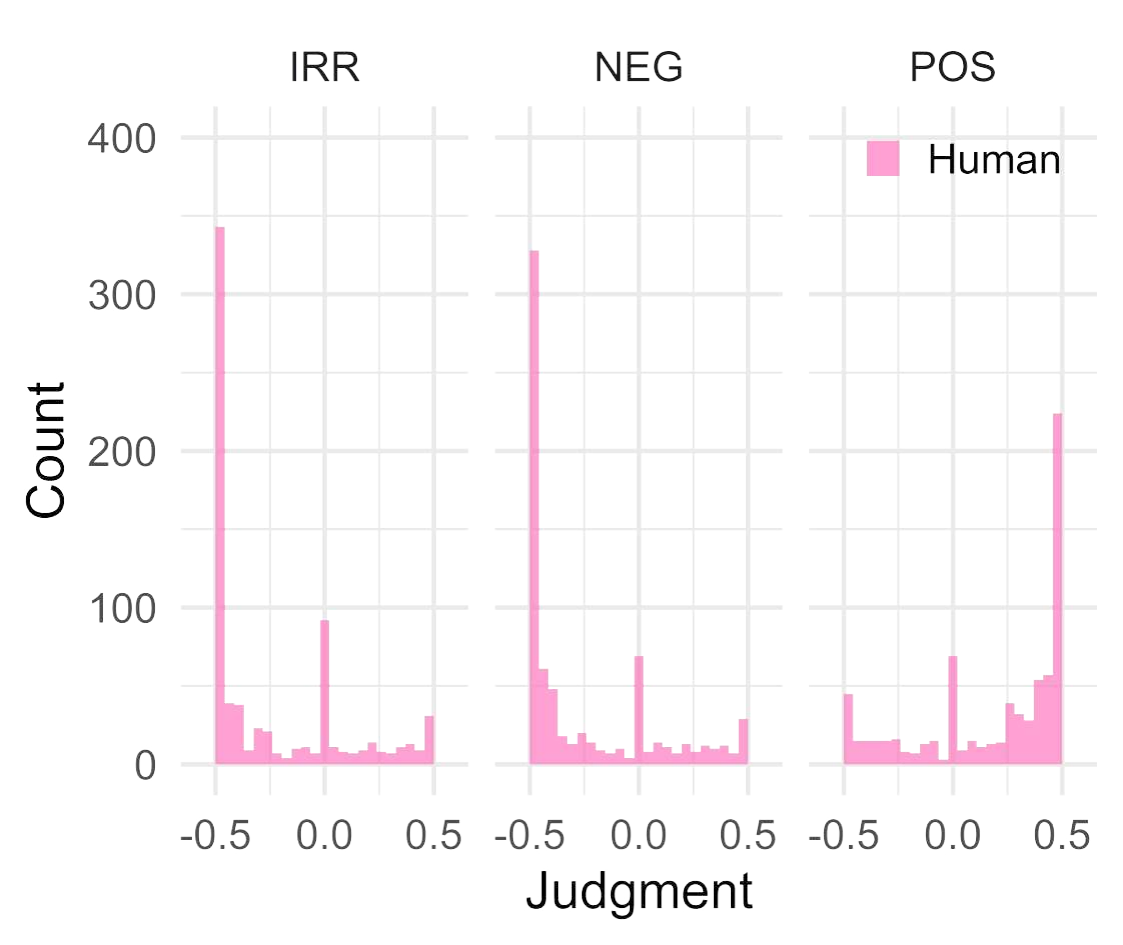}

  \caption{Distribution of center-scaled judgments across humans and LLMs. Top row: overall distributions. Bottom row: distributions divided by relation type. Left: Llama 70B (vanilla), center: Qwen 72B (vanilla), right: human data obtained from the study by~\citet{skovgaard2016relevance}.}
  \label{fig:histogram_human_llms_combined}
\end{figure*}

\paragraph{Post-Processing}
To robustly extract LLM judgments, we use a regular expression to match numbers between $0$ and $100$ following either the token “answer” or “assistant”. If no valid number in the expected range is found, the response is recorded as missing (\emph{NaN}). Notably, across all LLMs and prompting techniques, we report no missing responses. For subsequent analyses, all numerical ratings are mean-centered, resulting in a scale from $-0.5$ to $0.5$; all reported values reflect this transformation. \emph{p}-values (statistical significance: $p<0.05$) are reported to three decimal places, except when highly significant, in which case thresholds are used ($p < 0.001$, $p < 0.0001$).

\subsection{Prompts \& Models}\label{prompts_models}
We aim to examine the impact of model family, model size, and prompting technique on the conditional acceptability judgments of LLMs. Therefore, we assess a total of four language models: \texttt{Llama-3.1-8B-Instruct},\footnotemark{} \texttt{Llama-3.1-70B-\ Instruct},\footnotemark[\value{footnote}]\footnotetext{https://huggingface.co/meta-llama/models} \texttt{Qwen2.5-7B-Instruct},\footnotemark{} and \texttt{Qwen\
2.5-72B-Instruct},\footnotemark[\value{footnote}]\footnotetext{https://huggingface.co/Qwen/collections} using publicly accessible weights obtained from the Hugging Face platform. All models are prompted using a vanilla (zero-shot) and three-shot format, while the two larger models (Llama 70B and Qwen 72B) are also tested with chain-of-thought (CoT) prompting~\citep{NEURIPS2022_9d560961}.
For further details on models and prompts, please refer to Appendix~\ref{app_models_prompts}. Our code is publicly available at: 
\href{https://github.com/mainlp/conditional-acceptability}{{https://github.com/mainlp/conditional-acceptability}}.

\begin{table}[tbp]
\centering
\small
\begin{tabular}{lccc}
\toprule
 & \textbf{Cond. Prob.} & \textbf{If-Prob.} & \textbf{Acceptability} \\
\midrule
\textbf{Vanilla} &  &  &  \\
\midrule
Llama 8B & 0.7685 & 0.7724 & 0.7529 \\
Llama 70B & 0.9468 & 0.9344 & 0.9201 \\
Qwen 7B & 0.9768 & 0.9716 & 0.9450 \\
Qwen 72B & 0.9693 & 0.9673 & 0.9784 \\
\midrule
\textbf{Few-shot} &  &  &  \\
\midrule
Llama 8B & 0.7930 & 0.7538 & 0.7596 \\
Llama 70B & 0.9438 & 0.9556 & 0.8820 \\
Qwen 7B & 0.9542 & 0.9556 & 0.9596 \\
Qwen 72B & 0.9559 & 0.9437 & 0.9768 \\
\midrule
\textbf{CoT} &  &  &  \\
\midrule
Llama 70B & 0.9014 & 0.8497 & 0.8757 \\
Qwen 72B & 0.9331 & 0.9127 & 0.9661 \\
\bottomrule
\end{tabular}
\caption{Intraclass correlation scores across five model runs for each judgment type, model, and prompt type.}
\label{tab_rating_consistency}
\end{table}

\paragraph{Consistency of Model Responses}
We generate model responses via nucleus sampling ($\text{top-p} = 1.0$) and a temperature of $T = 1.0$. As this decoding procedure is non-deterministic, we ensure the consistency of the models' numerical ratings through repeated trials and a post-hoc correlation analysis. Each model is prompted five times per data sample to account for potential instabilities in ratings. By computing the intraclass correlation of these five ratings per sample, we further evaluate the reliability of the models' judgments.
As shown in Table~\ref{tab_rating_consistency}, all models demonstrate good-to-excellent consistency across all judgment types and prompting strategies.

\subsection{Statistical Methods} \label{method_statistical}
We follow the core analytic approach of~\citet{skovgaard2016relevance}, applying \emph{linear mixed-effects models} to investigate how conditional probability and relation type affect acceptability judgments of conditional statements.

\paragraph{Linear Mixed-Effects Models}
Mixed-effects models allow for the analysis of how conditional probability and relation type influence conditional acceptability judgments while accounting for additional sources of variation. In particular, different \emph{scenarios}, \emph{participants},\footnote{While we do not collect human data ourselves, we use human data from~\citet{skovgaard2016relevance} to compare LLM trends.} and LLM \emph{sampling iterations} introduce underlying variation that might otherwise obscure meaningful effects.
Linear mixed-effects models account for this structure by incorporating both fixed effects (the predictors of interest) and random effects (uncontrolled variation across groups)~\citep{brown2021introduction}.
For further information on model fitting, please refer to Appendix~\ref{app_linear_model}.

\paragraph{Additional Measures}
We report Type III ANOVA, fixed effect estimates, slopes, and estimated marginal means (EMMs) to assess the impact of predictors, such as conditional probability and relation types, on judgments of conditional statements. ANOVA tests show whether mean differences are statistically significant across conditions (e.g., relation types)~\citep{brown2021introduction, shaw1993anova, tabachnick2007experimental}, while EMMs represent the model-adjusted average outcomes for each condition, allowing for comparisons between groups while controlling for other variables~\citep{searle1980population, cranemmeanspackage, cranemmeansbasics}.

\section{Results}
Figure~\ref{fig:histogram_human_llms_combined} shows the distribution of vanilla LLM judgments (left: Llama 70B, center: Qwen 72B). Responses cluster around specific values, leading to a scattered overall distribution. This pattern is even more pronounced when split by relation type (bottom row); for example, Llama 70B gives almost exclusively low ratings for negative and irrelevant conditions. This supports prior findings that models prefer specific tokens or rating values in their judgments \citep{stureborg2024large, zheng2023large}.

\begin{figure}[tbp]
  \includegraphics[width=\columnwidth]{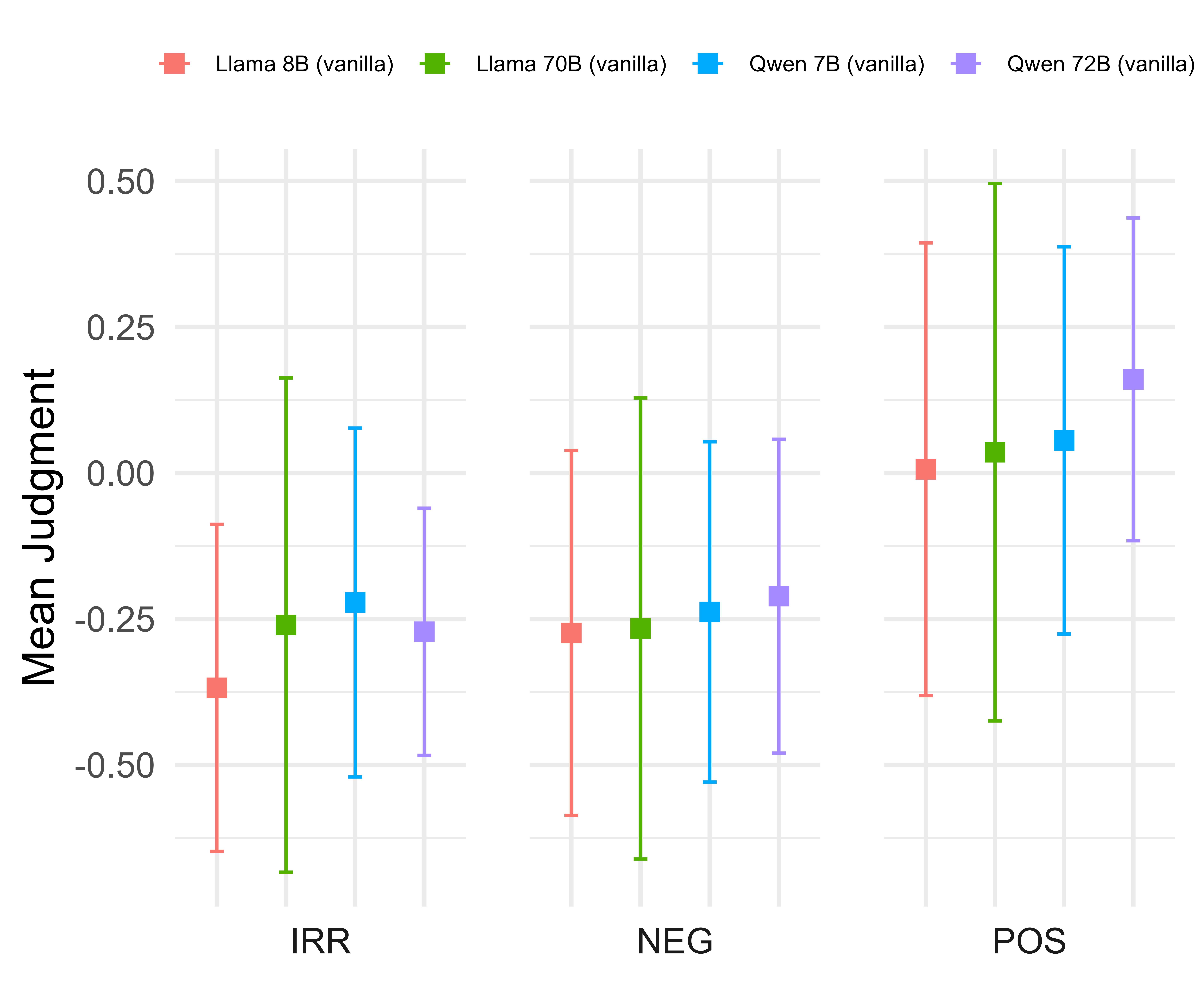}
  \caption{Mean and standard deviation across relevance conditions and vanilla models (red: Llama 8B, green: Llama 70B, blue: Qwen 7B, purple: Qwen 72B).}
  \label{plot:mean_sd_llms}
\end{figure}

\paragraph{Impact of Relation Type}
We find that all LLMs exhibit sensitivity to relevance when judging conditional statements. As shown in Figure~\ref{plot:mean_sd_llms}, mean judgments differ systematically across relation types: conditionals with a positive (supporting) relation receive more favorable ratings than those with a negative (contradicting) or irrelevant (unrelated) relation. The highest positive mean is observed for Qwen 72B (vanilla) at \num{0.1603125}, while Llama 8B (vanilla) shows the lowest, barely exceeding the midpoint, indicating that this model rates positively related statements less favorably than the others.
Notably, Llama 70B presents the greatest standard deviation of all vanilla models (see Figure~\ref{plot:mean_sd_llms}), which might be attributable to its spiked data distribution (mainly concentrating on either very high or very low values, see Figure~\ref{fig:histogram_human_llms_combined}).  Exact values for all models and prompting techniques can be found in Table~\ref{tab:general_means} in the Appendix.
ANOVA results confirm a highly significant main effect of relation type for all models except Llama 70B (vanilla) ($p =$ \pnum{0.06018}; see Table~\ref{tab:anova_Llama 70B (vanilla)} in Appendix~\ref{app_anova}).
Fixed effects estimates (Table~\ref{tab:fixed_effects_all}) further highlight these differences: for instance, Qwen 72B (vanilla) rates negatively related conditionals \num{0.203289611} higher, and positively related ones \num{0.303848052} higher, than irrelevant ones. In contrast, Llama 70B (vanilla) shows minimal differentiation between relation types (\num{0.07115076} for NEG, \num{0.01214315} for POS), suggesting lower sensitivity to relevance distinctions.
In summary, while most models exhibit significant sensitivity to semantic relation type, Llama 70B (vanilla) appears to do so less than other models.

\begin{table}[tbp]
  \small
  \centering
  \begin{tabular}{lcccc}
    \toprule
     & \textbf{$P(B \mid A)$} & \textbf{NEG} & \textbf{POS} & \textbf{Metric (P)} \\
    \midrule
    \textbf{Vanilla} &  &  &  & \\
    \midrule
    Llama 8B & \num{0.346609} & \num{0.09584093} & \num{0.29504530} & \num{-0.05664203} \\
    Llama 70B & \num{0.6523788} & \num{0.07115076} & \num{0.01214315} & \num{0.05375823} \\
    Qwen 7B & \num{0.384281} & \num{0.06443432} & \num{0.25591124} & \num{0.04571786} \\
    Qwen 72B & \num{0.2109758} & \num{0.203289611} & \num{0.303848052} & \num{-0.02578361} \\
    \midrule
    \textbf{Few-shot} &  &  &  & \\
    \midrule
    Llama 8B & \num{0.2946703} & \num{0.2327870} & \num{0.3277807} & \num{0.1207252} \\
    Llama 70B & \num{0.5080064} & \num{0.26198072} & \num{0.19393649} & \num{0.08352177} \\
    Qwen 7B & \num{0.4884646} & \num{0.12876907} & \num{0.23136938} & \num{0.05488655} \\
    Qwen 72B & \num{0.3009045} & \num{0.22520948} & \num{0.28394135} & \num{0.09276954} \\
    \midrule
    \textbf{CoT} &  &  &  & \\
    \midrule
    Llama 70B & \num{0.5065853} & \num{0.29857655} & \num{0.31324013} & \num{0.09614976} \\
    Qwen 72B & \num{0.1905983} & \num{0.27440384} & \num{0.34988825} & \num{0.06725216} \\
    \bottomrule
  \end{tabular}
  \caption{Fixed effect estimates for LLMs predicting probability and acceptability judgments. Effects are shown as contrasts relative to the reference levels: relation = irrelevant, metric = acceptability.}
  \label{tab:fixed_effects_all}
\end{table}

\begin{figure}[b!]
  \includegraphics[width=\columnwidth]{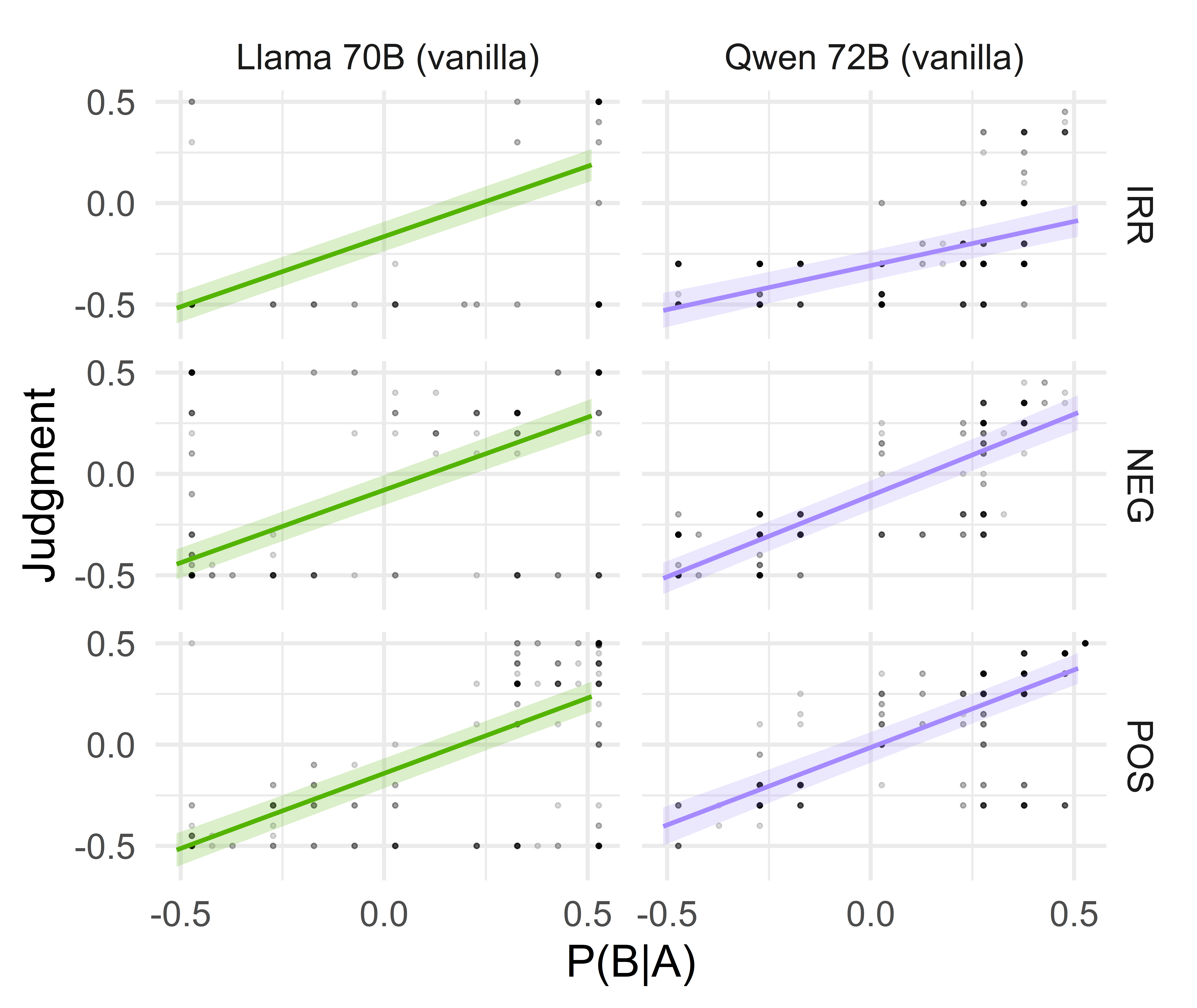}
  \caption{Scatterplots of data points and trend lines for Llama 70B (vanilla) and Qwen 72B (vanilla), divided by relation types. Plots for all vanilla models can be found in Figure~\ref{plot:scatter_trendlines_vanilla} in the Appendix.}
  \label{plot:scatter_trendlines}
\end{figure}

\paragraph{Impact of Conditional Probability}
For all LLMs, judgments of conditional statements are strongly influenced by the underlying conditional probability $P(B \mid A)$. ANOVA results (Tables~\ref{tab:anova_Llama 70B (vanilla)} to~\ref{tab:anova_Qwen 72B (CoT)} in the Appendix) confirm a highly significant main effect of conditional probability across all models ($p < 0.0001$). As shown in Figure~\ref{plot:scatter_trendlines}, judgments increase with rising conditional probability, a pattern further supported by fixed effects estimates in Table~\ref{tab:fixed_effects_all}, which quantify the effect of $P(B \mid A)$ on conditional statement judgments across models. Llama 70B (vanilla) shows the strongest sensitivity, with a slope of \num{0.6523788} per unit increase in $P(B \mid A)$, suggesting that, with the absence of relevance influence, it is particularly sensitive to conditional probability. Out of the vanilla models, Qwen 72B shows the weakest effect (\num{0.2109758}).

\begin{table*}[tbp]
  \small
  \centering
  \begin{tabular}{lrrrrrr}
    \toprule
     & \textbf{IRR $\times$ NEG} &  & \textbf{IRR $\times$ POS} &  & \textbf{NEG $\times$ POS} & \\
    \midrule
     & \textbf{Estimate} & \textbf{$p$} & \textbf{Estimate} & \textbf{$p$} & \textbf{Estimate} & \textbf{$p$}\\
    \midrule
    Llama 8B (vanilla) & \num{-0.101} & \pnum{0.0635} & \num{-0.277} & $<0.0001$ & \num{-0.176} & \pnum{0.0014} \\
    Llama 70B (vanilla) & \num{-0.0245} & \pnum{1.0000} & \num{-0.0573} & \pnum{0.6036} & \num{-0.0328} & \pnum{1.0000} \\
    Qwen 7B (vanilla) & \num{-0.1313} & \pnum{0.0020} & \num{-0.2222} & $<0.0001$ & \num{-0.0908} & \pnum{0.0292} \\
    Qwen 72B (vanilla) & \num{-0.3761} & $<0.0001$ & \num{-0.3466} & $<0.0001$ & \num{0.0295} & \pnum{0.4197} \\
    \midrule
    Llama 8B (few-shot) & \num{-0.2138} & $<0.001$ & \num{-0.1844} & \pnum{0.0008} & \num{0.0294} & \pnum{0.5725} \\
    Llama 70B (few-shot) & \num{-0.366} & $<0.0001$ & \num{-0.171} & $<0.001$ & \num{0.195} & $<0.001$ \\
    Qwen 7B (few-shot) & \num{-0.247} & $<0.0001$ & \num{-0.119} & \pnum{0.0166} & \num{0.127} & \pnum{0.0166} \\
    Qwen 72B (few-shot) & \num{-0.3503} & $<0.0001$ & \num{-0.2801} & $<0.0001$ & \num{0.0701} & \pnum{0.0768} \\
    \midrule
    Llama 70B (CoT) & \num{-0.3099} & $<0.0001$ & \num{-0.0945} & \pnum{0.0337} & \num{0.2154} & $<0.0001$ \\
    Qwen 72B (CoT) & \num{-0.5119} & $<0.0001$ & \num{-0.4365} & $<0.0001$ & \num{0.0754} & \pnum{0.0613} \\
    \bottomrule
  \end{tabular}
  \caption{Pairwise comparisons of interaction slopes for LLMs. SE: $0.04 - 0.05$. Statistical significance at $p < 0.05$.}
  \label{tab:condprob_slopes}
\end{table*}

\paragraph{How Conditional Probability and Relation Types Interact}
For all models except Llama 70B (vanilla), relation type significantly modulates the effect of conditional probability, as indicated by highly significant interaction effects in the ANOVA results ($p < 0.001$ for the vanilla models; see Tables~\ref{tab:anova_Llama 8B (vanilla)} to~\ref{tab:anova_Qwen 72B (vanilla)} in the Appendix). Table~\ref{tab:condprob_slopes} presents pairwise comparisons of interaction slopes, which indicate how much more strongly judgments increase with conditional probability for one relation type compared to another. These comparisons reveal distinct patterns across models.
Llama 70B (vanilla) shows no significant slope differences (IRR~$\times$~POS: $p=$ \pnum{0.6036}; others: $p=$ \pnum{1.0000}), reinforcing its insensitivity to relevance distinctions.
Among the remaining models, two distinct trends emerge. In the smaller models (Llama 8B and Qwen 7B, both vanilla), positive relations yield the steepest slopes (e.g., Llama 8B slope differences: IRR $\times$ POS: \num{-0.277}; NEG $\times$ POS: \num{-0.176}; $p \leq 0.001$ respectively), suggesting a stronger probability effect for supportive conditionals.
By contrast, larger models and all few-shot or CoT variants exhibit steeper slopes for negatively related conditionals than for positive ones (e.g., Llama 70B (few-shot): \num{0.195}, $p<0.001$) or, as in Qwen 72B (vanilla), show no significant difference between the two (\num{0.0295}, $p=$ \pnum{0.4197}). Not only does relevance modulate the effect of conditional probability, but the reverse also seems true: the influence of relation type depends on the underlying probability of the conditional statement. Pairwise comparisons of estimated marginal means at low, medium, and high probability levels (see Table~\ref{tab:emm} in the Appendix) reveal that in both Qwen models (7B and 72B, vanilla), all relation types differ significantly; except at low probability, where the irrelevant and negative conditions do not differ (e.g., Qwen 72B: $p = $ \pnum{0.2323}). A similar pattern appears in Llama 8B (vanilla), which shows no significant difference between negative and positive relations at low probability ($p=$ \pnum{0.1480}). These results suggest that while these LLMs reliably distinguish relation types at medium and high conditional probabilities, their sensitivity decreases at low probabilities; perhaps because relevance becomes harder to evaluate when the conditional itself seems implausible.

\paragraph{Impact of Judgment Type} \label{results_metric}
Up to this point, we have analyzed the if-probability and acceptability judgments of LLMs together. However, ANOVA results (see Tables~\ref{tab:anova_Llama 70B (vanilla)} to~\ref{tab:anova_Qwen 72B (CoT)} in the Appendix) suggest that some LLMs may differentiate between these two metrics. Only Llama 70B (vanilla) and Qwen 72B (vanilla) show no significant main effect of judgment type, though both approach the threshold of significance (Llama $p =$ \pnum{0.05767} Table~\ref{tab:anova_Llama 70B (vanilla)}; Qwen $p =$ \pnum{0.0682523} Table~\ref{tab:anova_Qwen 72B (vanilla)}).
Fixed effects estimates (using acceptability as the reference level, see Table~\ref{tab:fixed_effects_all}) reveal no consistent trend among the vanilla models: Qwen 72B and Llama 8B give slightly lower ratings to if-probability (e.g., Qwen: \num{-0.02578361}), while Qwen 7B and Llama 70B show the opposite pattern (both: $0.05$).
In contrast, all few-shot and CoT variants assign higher ratings to if-probability than to acceptability (see Table~\ref{tab:fixed_effects_all}). This effect is most pronounced in Llama 8B (few-shot: \num{0.1207252}) and smallest in Qwen 7B (few-shot: \num{0.05488655}). 
These results suggest that prompting examples consistently shift model ratings upward for probability judgments relative to acceptability. However, for zero-shot, this distinction appears less stable and may depend on model size or architecture. Larger models like Llama 70B and Qwen 72B appear less influenced by judgment type overall, whereas prompting reintroduces or amplifies this sensitivity.

\paragraph{Impact of Prompting Strategy} \label{results_prompting}
All LLMs appear highly susceptible to few-shot and CoT prompting, though the direction of these effects varies across models.
While few-shot prompting aids Llama 70B in incorporating relation types in its judgments (ANOVA: $p=$ \pnum{0.06018} to $< 0.0001$, see Tables~\ref{tab:anova_Llama 70B (vanilla)} and~\ref{tab:anova_Llama 70B (few-shot)} in the Appendix), it seems to introduce a negativity bias in other models:
As can be seen in Table~\ref{tab:condprob_slopes}, all few-shot variants produce steeper interaction slopes for negatively related conditionals than for positively related ones (e.g., Llama 70B (few-shot): \num{0.195},  $p<0.001$), or, as in Llama 8B (few-shot), eliminate the distinction entirely: $p = $ \pnum{0.5725}.
Prompting effects are also evident in the pairwise comparisons of estimated marginal means across conditional probability levels (see Table~\ref{tab:emm} in the Appendix). In the vanilla Qwen models, relation type distinctions collapse only at low probability between irrelevant and negative conditionals. However, the few-shot and CoT variants additionally produce insignificant differences between negative and positive conditionals at \emph{high} conditional probability (e.g., Qwen 72B CoT $p=$ \pnum{1.0000}), suggesting that the prompting examples may, in some cases, reduce sensitivity to fine-grained relevance contrasts. Contrary to expectations, prompting examples do not appear to increase alignment with human behavior, which typically shows great sensitivity to relevance cues. Notably, the Llama models do not exhibit this degradation under prompting, except for Llama 70B CoT, which converges negative and positive at \emph{medium} conditional probability ($p=$ \pnum{0.4593}).

\paragraph{Impact of Model Size}
Intriguingly, smaller models show a more consistent influence of relation type than their larger counterparts.
For instance, as shown in Table~\ref{tab:condprob_slopes}, Llama 8B (vanilla) exhibits significant slope differences between relation types and strong fixed effects estimates, unlike Llama 70B (vanilla), which shows virtually no influence of relevance. Similarly, Qwen 7B (vanilla) does not display the reduced contrast between positive and negative slopes observed in Qwen 72B (vanilla).
This pattern suggests that smaller models may more consistently integrate relevance cues, especially for supporting conditionals.
Conversely, larger models tend to show reduced variation across the type of judgment (if-probability vs. acceptability), indicating a different kind of internal consistency.

\subsection{Comparison Humans vs. LLMs} \label{comparison_humans_llms}

\paragraph{Data Distribution}
Overall, the rating patterns of humans and LLMs show broad similarities, yet systematic differences emerge. As shown in Figure~\ref{fig:histogram_human_llms_combined}, the general distribution of human ratings appears continuous, whereas LLMs tend to concentrate responses around specific values.
Broken down by relation type (bottom row), both humans and LLMs generally rate positively related conditionals more favorably, and irrelevant and negative conditionals less so. However, human ratings show tighter clustering: irrelevant and negative statements consistently fall at the lower end of the scale, while positive ones cluster at the higher end. In contrast, LLMs exhibit greater dispersion, with outlier judgments appearing across the scale. For example, Llama 70B (vanilla) assigns both very high and very low ratings to positively related conditionals.

\paragraph{Systematic Divergences}
Despite these differences, the overall mean ratings by relation type are similar. Human means for the irrelevant and negative conditions (IRR: \num{-0.2505191}, NEG: \num{-0.2625273}) closely match the ones of all LLMs, which range from \num{-0.37} to \num{-0.20} for IRR and \num{-0.27} to \num{-0.19} for NEG (see Table~\ref{tab:general_means} in the Appendix). The biggest difference occurs in the positive condition (humans: \num{0.1922131}), where most LLMs fall short. 
Both Llama models and the smaller Qwen variant barely exceed the midpoint (Llama 8B: $0.01$, 70B: $0.04$; Qwen: $0.06$).
Qwen 72B is the only LLM that approximates this high mean across prompting techniques (vanilla: \num{0.1603125}, few-shot: \num{0.1521458}, CoT: \num{0.1393958}).
In terms of the influence of predictors, both humans and LLMs show strong effects of conditional probability and relation type. For humans, ANOVA results (see Table~\ref{tab:anova_Human} in the Appendix) indicate significant main effects of probability and relevance, as well as their interaction ($p < 0.0001$), but no effect of judgment metric. This stands in contrast to LLMs, where the smaller models and all few-shot and CoT models consistently show modest but significant effects of the judgment type.
In terms of slope comparisons, human pairwise interactions across relation types are highly significant ($p < 0.0001$), indicating robust modulation by relevance. This pattern most closely resembles Qwen 7B (vanilla), Llama 70B (few-shot), and Qwen 7B (few-shot), although the LLMs generally produce weaker effects and differ in direction: humans show steeper slopes for positively related conditionals, whereas LLMs sometimes favor negative ones. Finally, estimated marginal means comparisons show that humans distinguish between relation types at all levels of conditional probability; except between irrelevant and negative conditionals at low probability ($p < 0.0001$ for all other contrasts, see Table~\ref{tab:emm}). This pattern is mirrored in the Qwen models and partially in the Llama models, which also exhibit collapsed distinctions at low probability levels.

In sum, while all LLMs except Llama 70B (vanilla) incorporate both conditional probability $P(B \mid A)$ and semantic relevance in their judgments of conditionals, they do not exhibit the systematic integration of these factors observed in humans. In particular, the consistently elevated ratings and steepest interaction slopes for positively related conditionals appear to reflect a uniquely human sensitivity that LLMs can only approximate.

\begin{table}[tbp]
\centering
\small
\renewcommand{\arraystretch}{1.15}
\setlength{\tabcolsep}{6pt}

\begin{tabularx}{\linewidth}{
  >{\raggedright\arraybackslash}p{0.26\linewidth}
  *{4}{>{\centering\arraybackslash}X}
}
\toprule
& \multicolumn{2}{c}{\textbf{Sentence Prob.}} & \multicolumn{2}{c}{\textbf{Perplexity}} \\
\cmidrule(lr){2-3} \cmidrule(lr){4-5}
 & $\mathbf{r}$ & $\mathbf{\rho}$ & $\mathbf{r}$ & $\mathbf{\rho}$ \\
\midrule
Cond. Prob. & & & & \\
\midrule
Llama 8B & 0.15 & 0.25 & -0.28 & -0.32 \\
Llama 70B & 0.16 & 0.26 & -0.28 & -0.38 \\
Qwen 7B & 0.08 & 0.22 & -0.17 & -0.27 \\
Qwen 72B & 0.16 & 0.33 & -0.21 & -0.36 \\
\midrule
If-Prob. & & & & \\
\midrule
Llama 8B & 0.19 & 0.25 & -0.32 & -0.32 \\
Llama 70B & 0.11 & 0.31 & -0.23 & -0.31 \\
Qwen 7B & 0.07 & 0.20 & -0.08 & -0.27 \\
Qwen 72B & 0.14 & 0.19 & -0.19 & -0.31 \\
\bottomrule
\end{tabularx}
\caption{Pearson $\left(\mathbf{r}\right)$ and Spearman correlation $\left(\mathbf{\rho}\right)$ of sentence probability and perplexity with probability judgments obtained through direct elicitation.}
\label{tab_correlation_sentenceprob}
\end{table}

\subsection{Eliciting Probability Judgments} \label{elicitation_methods_main}
In our study, we directly prompt models to judge the probability or acceptability of a conditional statement. While this method is well-established and has been used in existing studies~\citep{tian-etal-2023-just, yang2025on}, there also exist other methods for eliciting ratings from LLMs.\footnote{A discussion of direct elicitation and alternative methods is provided in Appendix~\ref{app_elicitation_methods}.} For instance, we can compute a model’s sentence probability or its perplexity as an estimate of the probability it assigns to a given statement.

To compare sentence probabilities with our direct elicitation approach, we compute the model-assigned probabilities of both the conditional probability $P(B \mid A)$ and the if-probability $P(\text{``If A, then B''})$. In particular, for conditional probability, we use the following prompt $M$:

\begin{tcolorbox}[
    colback=MidnightBlue!5!white,
    colframe=MidnightBlue,
    boxrule=0.4pt,
    rounded corners,
    width=\columnwidth,
    before upper={\parindent0pt}
  ]
    Context: <context>\\
    Given: <A> \\
    Consequence:
\end{tcolorbox}

and compute $P(B \mid M)$ using the probabilities the model assigns to each token in $B$. Similarly, for if-probability, we compute $P(\text{If A, then B} \mid M)$ using the prompt $M$:

\begin{tcolorbox}[
    colback=MidnightBlue!5!white,
    colframe=MidnightBlue,
    boxrule=0.4pt,
    rounded corners,
    width=\columnwidth,
    before upper={\parindent0pt}
  ]
    Context: <context>\\
    Statement:
\end{tcolorbox}

Because sentence probability strongly depends on sentence length, we additionally compute the model's perplexity over the statements. To study how well these probability-based judgments correlate with those obtained via direct elicitation, we compute both Pearson and Spearman correlations for each metric. As shown in Table~\ref{tab_correlation_sentenceprob}, we observe a weak positive correlation for sentence probability and a weak-to-moderate negative correlation for perplexity scores, underlining that these metrics are related but ultimately capture different aspects.

\section{Related Work}

\paragraph{Conditional Reasoning in LLMs}
Several recent studies examine LLMs' abilities in conditional and causal reasoning, though without addressing the acceptability of statements themselves.
\citet{liu2023magic} find that Code-LLMs outperform text-only models on abductive and counterfactual tasks when prompted in code, suggesting improved causal sensitivity.
\citet{dettki2025large} show that LLMs draw on domain knowledge in causal inference rather than relying purely on abstract rules.
Other work focuses on logical inference:~\citet{holliday2024conditional} report frequent errors in detecting invalid inferences, though CoT prompting improves performance.~\citet{debray2023conditional} find that models like OPT show limited validity assessment but some semantic sensitivity when compared to human judgments.

\section{Conclusion}
In this paper, we investigate how LLMs assess conditional statements, a task that has long been debated in the context of human reasoning, but remains underexplored for LLMs. Specifically, we examine whether LLMs rely on conditional probability, relation type, or a combination of both when judging the acceptability of conditionals.
Our findings suggest that both factors contribute to LLM judgments, although the degree of this influence varies substantially across models and prompting strategies.
Notably, while LLMs and humans perform generally similar, key distinctions emerge, suggesting that LLMs may not \emph{consistently} integrate both semantic relation and probabilistic strength in the systematic way that humans do.

\section{Limitations}
While our work sheds light on LLM capabilities in assessing conditional statements, several limitations warrant consideration and future investigation.

\paragraph{Human Data}
As discussed in Appendix~\ref{app_human_study}, the human judgment data from~\citet{skovgaard2016relevance} is collected via crowdsourcing. Although quality controls are implemented (such as restricting participation to native English speakers and setting time thresholds), these precautions are introduced only after data collection had already taken place. Consequently, the dataset features uneven response counts per conditional statement (e.g., some statements rated by ten participants, others by only three), introducing variability and inconsistency.
Future work could benefit from eliciting human judgments in more controlled environments, such as laboratory or classroom settings, ideally synchronized with LLM data collection to enable direct comparisons under matched conditions.

\paragraph{Prompts and Prompting}
As noted in Section~\ref{results_prompting}, the few-shot and CoT prompting conditions appear to bias model sensitivity towards negatively related conditionals. Since this effect occurred in all few-shot and CoT variants, the random selection of prompt examples (see Appendix~\ref{app_fewshot_cot}) may have inadvertently biased model behavior.
Given the well-documented prompt sensitivity of LLMs~\citep{errica2024did, loya2023exploring, sclar2023quantifying}, future work should take greater care in constructing prompt sets. One possible improvement could be to include a more balanced set of few-shot examples, e.g., one for each combination of relation type and conditional probability level (resulting in nine examples total). This might help ensure that models are not inadvertently primed to favor particular judgment patterns.

\paragraph{Limits of Relying Solely on Ratings}
A key limitation of this study is the reliance on numerical ratings alone. While this approach enables systematic comparisons, it may not fully capture the reasoning processes that underlie those judgments. Conditional evaluation is a complex task for both humans and LLMs. That some aspects elicit greater concordance across models and participants than others suggests that numerical ratings may obscure important distinctions in underlying reasoning. 
Future work may benefit from incorporating richer diagnostic measures. A more informative approach could involve qualitative analysis of both LLM and human reasoning. Such methods would offer deeper insight into the reasoning strategies underlying judgment patterns and help determine whether similar ratings reflect genuinely similar reasoning or merely coincidental alignment. This, in turn, would enable more nuanced evaluations of LLM–human alignment in conditional reasoning.

\section*{Acknowledgments}
We would like to thank Alessandro Bogani for his initial input to the project and for his help in identifying relevant literature. Our appreciation extends to the anonymous reviewers for their valuable comments and suggestions. Lastly, we acknowledge the support for BP through the ERC Consolidator Grant DIALECT 101043235.

\bibliography{custom}

@book{evans2004conditionalsIF,
  title={If: Supposition, Pragmatics, and Dual Processes},
  author={Evans, Jonathan St B T and Over, David E},
  year={2004},
  publisher={Oxford University Press}
}

@book{byrne2007imagination,
  title={The Rational Imagination: How People Create Alternatives to Reality},
  author={Byrne, Ruth M. J.},
  year={2007},
  publisher={MIT Press}
}

@book{douven2015epistemology,
  title={The epistemology of indicative conditionals: Formal and empirical approaches},
  author={Douven, Igor},
  year={2015},
  publisher={Cambridge University Press}
}

@book{vasishth2010foundations,
  title={The foundations of statistics: A simulation-based approach},
  author={Vasishth, Shravan and Broe, Michael},
  year={2010},
  publisher={Springer Science \& Business Media},
  doi={10.1007/978-3-642-16313-5}
}

@book{tabachnick2007experimental,
  title={Experimental designs using ANOVA},
  author={Tabachnick, Barbara G and Fidell, Linda S},
  volume={724},
  year={2007},
  publisher={Thomson/Brooks/Cole Belmont, CA},
  url={https://www.researchgate.net/profile/Barbara-Tabachnick/publication/259465542_Experimental_Designs_Using_ANOVA/links/5e6bb05f92851c6ba70085db/Experimental-Designs-Using-ANOVA.pdf}
}

@book{hardman2003thinking,
  title={Thinking: psychological perspectives on reasoning, judgment and decision making},
  author={Hardman, David and Macchi, Laura},
  year={2003},
  publisher={Wiley Online Library},
  url={https://al-edu.com/wp-content/uploads/2014/05/HarmanMacchi-eds-Thinking-Psychological-Perspectives-on-Reasoning-Judgement-and-Decision-Making.pdf}
}

@article{barrowman2014correlation,
  title={Correlation, causation, and confusion},
  author={Barrowman, Nick},
  journal={The New Atlantis},
  pages={23--44},
  year={2014},
  publisher={JSTOR},
  url={https://www.jstor.org/stable/43551404}
}

@article{oaksford2003conditional,
  title={Conditional probability and the cognitive science of conditional reasoning},
  author={Oaksford, Mike and Chater, Nick},
  journal={Mind \& Language},
  volume={18},
  number={4},
  pages={359--379},
  year={2003},
  publisher={Wiley Online Library},
  doi={https://doi.org/10.1111/1468-0017.00232}
}

@article{berto2021indicative,
  title={Indicative conditionals: Probabilities and relevance},
  author={Berto, Francesco and {\"O}zg{\"u}n, Ayb{\"u}ke},
  journal={Philosophical Studies},
  volume={178},
  number={11},
  pages={3697--3730},
  year={2021},
  publisher={Springer},
  doi={https://doi.org/10.1007/s11098-021-01622-3}
}

@article{hao2024llm,
  title={Llm reasoners: New evaluation, library, and analysis of step-by-step reasoning with large language models},
  author={Hao, Shibo and Gu, Yi and Luo, Haotian and Liu, Tianyang and Shao, Xiyan and Wang, Xinyuan and Xie, Shuhua and Ma, Haodi and Samavedhi, Adithya and Gao, Qiyue and others},
  journal={arXiv preprint arXiv:2404.05221},
  year={2024},
  doi={https://doi.org/10.48550/arXiv.2404.05221}
}

@article{clark2020transformers,
  title={Transformers as soft reasoners over language},
  author={Clark, Peter and Tafjord, Oyvind and Richardson, Kyle},
  journal={arXiv preprint arXiv:2002.05867},
  year={2020},
  doi={https://doi.org/10.48550/arXiv.2002.05867}
}

@article{yu2024improving,
  title={Improving causal reasoning in large language models: A survey},
  author={Yu, Longxuan and Chen, Delin and Xiong, Siheng and Wu, Qingyang and Liu, Qingzhen and Li, Dawei and Chen, Zhikai and Liu, Xiaoze and Pan, Liangming},
  journal={arXiv preprint arXiv:2410.16676},
  year={2024},
  doi={https://doi.org/10.48550/arXiv.2410.16676}
}

@article{evans2003conditionals,
  title={Conditionals and conditional probability.},
  author={Evans, Jonathan St B T and Handley, Simon J and Over, David E},
  journal={Journal of Experimental Psychology: Learning, Memory, and Cognition},
  volume={29},
  number={2},
  pages={321--335},
  year={2003},
  publisher={American Psychological Association},
  doi={10.1037/0278-7393.29.2.321}
}

@article{douven2012indicatives,
  title={Indicatives, concessives, and evidential support},
  author={Douven, Igor and Verbrugge, Sara},
  journal={Thinking \& Reasoning},
  volume={18},
  number={4},
  pages={480--499},
  year={2012},
  publisher={Taylor \& Francis},
  doi={10.1080/13546783.2012.716009}
}

@inproceedings{mondorf-plank-2024-comparing,
    title = "Comparing Inferential Strategies of Humans and Large Language Models in Deductive Reasoning",
    author = "Mondorf, Philipp  and
      Plank, Barbara",
    editor = "Ku, Lun-Wei  and
      Martins, Andre  and
      Srikumar, Vivek",
    booktitle = "Proceedings of the 62nd Annual Meeting of the Association for Computational Linguistics (Volume 1: Long Papers)",
    month = aug,
    year = "2024",
    address = "Bangkok, Thailand",
    publisher = "Association for Computational Linguistics",
    url = "https://aclanthology.org/2024.acl-long.508/",
    doi = "10.18653/v1/2024.acl-long.508",
    pages = "9370--9402"
}

@inproceedings{tian-etal-2023-just,
    title = "Just Ask for Calibration: Strategies for Eliciting Calibrated Confidence Scores from Language Models Fine-Tuned with Human Feedback",
    author = "Tian, Katherine  and
      Mitchell, Eric  and
      Zhou, Allan  and
      Sharma, Archit  and
      Rafailov, Rafael  and
      Yao, Huaxiu  and
      Finn, Chelsea  and
      Manning, Christopher",
    editor = "Bouamor, Houda  and
      Pino, Juan  and
      Bali, Kalika",
    booktitle = "Proceedings of the 2023 Conference on Empirical Methods in Natural Language Processing",
    month = dec,
    year = "2023",
    address = "Singapore",
    publisher = "Association for Computational Linguistics",
    url = "https://aclanthology.org/2023.emnlp-main.330/",
    doi = "10.18653/v1/2023.emnlp-main.330",
    pages = "5433--5442"
}

@inproceedings{
yang2025on,
title={On Verbalized Confidence Scores for {LLM}s},
author={Daniel Yang and Yao-Hung Hubert Tsai and Makoto Yamada},
booktitle={ICLR Workshop: Quantify Uncertainty and Hallucination in Foundation Models: The Next Frontier in Reliable AI},
year={2025},
url={https://openreview.net/forum?id=CVRdNQvFPE}
}

@article{over2007probability,
  title={The probability of causal conditionals},
  author={Over, David E and Hadjichristidis, Constantinos and Evans, Jonathan St B T and Handley, Simon J and Sloman, Steven A},
  journal={Cognitive psychology},
  volume={54},
  number={1},
  pages={62--97},
  year={2007},
  publisher={Elsevier},
  doi={10.1016/j.cogpsych.2006.05.002}
}

@article{skovgaard2016relevance,
  title={The relevance effect and conditionals},
  author={Skovgaard-Olsen, Niels and Singmann, Henrik and Klauer, Karl Christoph},
  journal={Cognition},
  volume={150},
  pages={26--36},
  year={2016},
  publisher={Elsevier},
  doi={http://dx.doi.org/10.1016/j.cognition.2015.12.017}
}

@article{dettki2025large,
  title={Do Large Language Models Reason Causally Like Us? Even Better?},
  author={Dettki, Hanna M and Lake, Brenden M and Wu, Charley M and Rehder, Bob},
  journal={arXiv preprint arXiv:2502.10215},
  year={2025},
  doi={https://doi.org/10.48550/arXiv.2502.10215}
}

@article{liu2023magic,
  title={The magic of IF: Investigating causal reasoning abilities in large language models of code},
  author={Liu, Xiao and Yin, Da and Zhang, Chen and Feng, Yansong and Zhao, Dongyan},
  journal={arXiv preprint arXiv:2305.19213},
  year={2023},
  doi={https://doi.org/10.48550/arXiv.2305.19213}
}

@article{holliday2024conditional,
  title={Conditional and modal reasoning in large language models},
  author={Holliday, Wesley H and Mandelkern, Matthew and Zhang, Cedegao E},
  journal={arXiv preprint arXiv:2401.17169},
  year={2024},
  doi={https://doi.org/10.48550/arXiv.2401.17169}
}

@phdthesis{debray2023conditional,
  title={Conditional Reasoning in Large Language Models},
  author={Debray, Samuel},
  year={2023},
  school={Universiteit van Amsterdam},
  url={https://perso.crans.org/sdebray/files/ARPEReport.pdf}
}

@article{brown2021introduction,
  title={An introduction to linear mixed-effects modeling in R},
  author={Brown, Violet A},
  journal={Advances in Methods and Practices in Psychological Science},
  volume={4},
  number={1},
  year={2021},
  publisher={Sage Publications Sage CA: Los Angeles, CA},
  doi={10.1177/2515245920960351}
}

@article{shaw1993anova,
  title={Anova for Unbalanced Data: An Overview},
  author={Shaw, Ruth G and Mitchell-Olds, Thomas},
  journal={Ecology},
  volume={74},
  number={6},
  pages={1638--1645},
  year={1993},
  url={https://www.researchgate.net/profile/Witold-Orlik/post/How_to_handle_missing_values_in_two_way_repeated_measures_ANOVA/attachment/59d62d1779197b807798b5bb/AS%3A348966169399296%401460211426860/download/unbalPaper.pdf}
}

@article{bates2015fitting,
  title={Fitting linear mixed-effects models using lme4},
  author={Bates, Douglas and M{\"a}chler, Martin and Bolker, Ben and Walker, Steve},
  journal={Journal of statistical software},
  volume={67},
  pages={1--48},
  year={2015},
  doi={10.18637/jss.v067.i01}
}

@article{searle1980population,
  title={Population Marginal Means in the Linear Model: An Alternative to Least Squares Means},
  author={Searle, SR and Speed, FM and Milliken, GA},
  journal={American Statistician},
  pages={216--221},
  year={1980},
  publisher={JSTOR},
  doi={https://doi.org/10.2307/2684063}
}

@article{sclar2023quantifying,
  title={Quantifying Language Models' Sensitivity to Spurious Features in Prompt Design or: How I learned to start worrying about prompt formatting},
  author={Sclar, Melanie and Choi, Yejin and Tsvetkov, Yulia and Suhr, Alane},
  journal={arXiv preprint arXiv:2310.11324},
  year={2023},
  doi={https://doi.org/10.48550/arXiv.2310.11324}
}

@article{loya2023exploring,
  title={Exploring the Sensitivity of LLMs' Decision-Making Capabilities: Insights from Prompt Variation and Hyperparameters},
  author={Loya, Manikanta and Sinha, Divya Anand and Futrell, Richard},
  journal={arXiv preprint arXiv:2312.17476},
  year={2023},
  doi={https://doi.org/10.48550/arXiv.2312.17476}
}

@article{errica2024did,
  title={What did I do wrong? quantifying LLMs' sensitivity and consistency to prompt engineering},
  author={Errica, Federico and Siracusano, Giuseppe and Sanvito, Davide and Bifulco, Roberto},
  journal={arXiv preprint arXiv:2406.12334},
  year={2024},
  doi={https://doi.org/10.48550/arXiv.2406.12334}
}

@misc{cranemmeansbasics,
    author =  {{CRAN}},
    title = {Quick start guide for emmeans},
    year = {2025},
    url = {https://cran.r-project.org/web/packages/emmeans/vignettes/basics.html},
    note = {Accessed: 2025-05-30}
}

@misc{cranemmeanspackage,
    author =  {{CRAN}},
    title = {Package ‘emmeans’},
    year = {2025},
    url = {https://cran.r-project.org/web/packages/emmeans/emmeans.pdf},
    note = {Accessed: 2025-05-30}
}

@inproceedings{
mondorf2024beyond,
title={Beyond Accuracy: Evaluating the Reasoning Behavior of Large Language Models - A Survey},
author={Philipp Mondorf and Barbara Plank},
booktitle={First Conference on Language Modeling},
year={2024},
url={https://openreview.net/forum?id=Lmjgl2n11u}
}

@article{stureborg2024large,
  title={Large language models are inconsistent and biased evaluators},
  author={Stureborg, Rickard and Alikaniotis, Dimitris and Suhara, Yoshi},
  journal={arXiv preprint arXiv:2405.01724},
  year={2024},
  doi = {https://doi.org/10.48550/arXiv.2405.01724}
}

@article{zheng2023large,
  title={Large language models are not robust multiple choice selectors},
  author={Zheng, Chujie and Zhou, Hao and Meng, Fandong and Zhou, Jie and Huang, Minlie},
  journal={arXiv preprint arXiv:2309.03882},
  year={2023},
  doi = {https://doi.org/10.48550/arXiv.2309.03882}
}

@inproceedings{NEURIPS2022_9d560961,
 author = {Wei, Jason and Wang, Xuezhi and Schuurmans, Dale and Bosma, Maarten and ichter, brian and Xia, Fei and Chi, Ed and Le, Quoc V and Zhou, Denny},
 booktitle = {Advances in Neural Information Processing Systems},
 editor = {S. Koyejo and S. Mohamed and A. Agarwal and D. Belgrave and K. Cho and A. Oh},
 pages = {24824--24837},
 publisher = {Curran Associates, Inc.},
 title = {Chain-of-Thought Prompting Elicits Reasoning in Large Language Models},
 url = {https://proceedings.neurips.cc/paper_files/paper/2022/file/9d5609613524ecf4f15af0f7b31abca4-Paper-Conference.pdf},
 volume = {35},
 year = {2022}
}

\appendix

\begin{table*}[t!]
  \small
  \centering
  \begin{tabular}{lcccc}
    \toprule
    \textbf{Relation / Probability} & \textbf{High–High} & \textbf{High–Low} & \textbf{Low–High} & \textbf{Low–Low} \\
    \midrule
    \textbf{Positive (POS)} & POS $\times$ High–High & POS $\times$ High–Low & POS $\times$ Low–High & POS $\times$ Low–Low \\
    \textbf{Negative (NEG)} & NEG $\times$ High–High & NEG $\times$ High–Low & NEG $\times$ Low–High & NEG $\times$ Low–Low \\
    \textbf{Irrelevant (IRR)} & IRR $\times$ High–High & IRR $\times$ High–Low & IRR $\times$ Low–High & IRR $\times$ Low–Low \\
    \bottomrule
  \end{tabular}
  \caption{Design matrix showing all 12 conditional statement configurations for a single scenario, based on 3 relation types (positive, negative, irrelevant) and 4 prior probability levels (high-high, high-low, low-high, low-low).}
  \label{tab:config_matrix}
\end{table*}

\section{Appendix}
\label{sec:appendix}

\subsection{Prior Human Study} \label{app_human_study}

\paragraph{Dataset} \label{app_dataset}
In the dataset, as shown in Table~\ref{tab:config_matrix}, each conditional varies along two dimensions: The first dimension is prior probability assignment, where antecedent and consequent are each classified by ~\citet{skovgaard2016relevance} as either high or low probability based on contextual cues provided in the scenario. This yields four prior configurations (\emph{high-high, high-low, low-high, low-low}). According to ~\citet{skovgaard2016relevance}, this manipulation aims not to test probability effects directly, but to increase the spread of conditional probability judgments across statements.

The second dimension is the relation type, indicating the type of connection between antecedent $A$ and consequent $B$. A positive ($POS$) relation means $A$ supports $B$, a negative ($NEG$) relation means $A$ undermines $B$, and an irrelevant ($IRR$) relation means $A$ is unrelated to $B$.

In total, this then results in 12 statements per scenario, each representing a unique combination of prior probability and relation type.

\paragraph{Data Collection}
\citet{skovgaard2016relevance} use this experimental setup to investigate how humans assess the acceptability of conditional statements \emph{If $A$, then $B$}. Their dataset and supplementary materials are openly available on \href{https://osf.io/j4swp/}{OSF}. All ratings are collected via a crowd-sourcing platform (\emph{CrowdFlower}), with participants from the United States, United Kingdom, and Australia. With this setup, the study shows that human judgments are sensitive to the semantic relation between $A$ and $B$ (positive, negative, or irrelevant) as well as the conditional probability of conditional statements.

While appropriate measures are implemented to ensure data quality, such as restricting participation to native English speakers and setting minimum and maximum completion times, these precautions are introduced only after data collection had already taken place. As a result, the dataset contains variable numbers of responses per conditional statement (e.g., some statements are rated by ten participants, others by only three), which introduces inconsistency.

\subsection{Models and Prompts} \label{app_models_prompts}

\paragraph{Language Models}

This study employs four publicly available large language models of comparable size ($7-8$ and $70-72$ billion parameters): Llama 3.1 (8B and 70B) and Qwen 2.5 (7B and 72B). All models are instruction-tuned variants designed to follow user prompts and generate coherent natural language responses.

A summary of key model specifications is provided in Table~\ref{table_llm_description_overview}. The Llama models, released in July 2024, consist of either 8.03 billion or 70.6 billion parameters, and the Qwen models, released in September 2024, contain 7.61 billion or 72.7 billion parameters. All models accommodate context windows of up to 128,000 tokens.

\begin{table}[bp]
    \centering
    \small
    \begin{tabular}{lcccc}
        \toprule
        \textbf{Model ID} & \textbf{Parameters} & \textbf{Context Length} \\
        \midrule
        Llama-3.1-8B-it & 8.03B & 128K \\
        Llama-3.1-70B-it & 70.6B & 128K \\
        Qwen2.5-7B-it & 7.61B & 128K \\
        Qwen2.5-72B-it & 72.7B & 128K \\
        \bottomrule
    \end{tabular}
    \caption{Overview of selected large language models used in this study.}
    \label{table_llm_description_overview}
\end{table}

\paragraph{Prompts}
While the vanilla prompts follow a zero-shot format, providing only the task instruction, the scenario context, and the input item, the few-shot variants additionally include three example judgments based on a shared narrative context (see Figure~\ref{fig_llm_prompt_fewshot_examples}). These examples span a range of output values (high, medium, low) to illustrate the rating scale. The same context and examples are used across all judgment types, with minor numerical variation to avoid repetition. Relation types are omitted to avoid bias.
All prompts are constructed using each model’s respective chat formatting template. Instruction-tuned LLMs typically expect input messages to be wrapped in a specific format. These templates are applied using each model’s tokenizer. Where applicable, the model’s default system prompt is used. Model outputs are generated using the default decoding parameters provided by the Hugging Face Transformers library, with nucleus sampling ($top\_p = 1.0$) and temperature ($T = 1.0$).

\paragraph{Outputs}
Each model judgment is expected to be a numerical value between $0$ (completely improbable or unacceptable) and $100$ (highly probable or acceptable). To elicit consistent responses, all prompts explicitly instruct models to return a single numerical value using both a textual instruction (\texttt{Respond only with this number and nothing else.}) and a predefined output format (\texttt{Answer: <number>}).

\subsection{Prompt Templates} \label{app_prompt_templates}

\paragraph{Vanilla}

Figure~\ref{fig_llm_prompt_vanilla_cprob} shows the vanilla prompt used to elicit \emph{conditional probability} estimates. Similar to the original human study~\citep{skovgaard2016relevance}, the model is given a context, a suppositional assumption (e.g., “Suppose Nicole sunbathes on the beach.”), and a target sentence (e.g., “She will get sunburned.”). It is then asked how probable the target sentence is, given the assumption.

Figure~\ref{fig_llm_prompt_vanilla_ifprob} shows the vanilla prompt used to elicit \emph{if-probability} judgments. In this setup, the model is shown a context and a conditional statement (e.g., “If Nicole sunbathes on the beach, then she will get sunburned.”) and asked how probable the conditional is.

Figure~\ref{fig_llm_prompt_vanilla_ifacc} presents the vanilla prompt for \emph{acceptability} ratings. The model receives a context and a conditional statement and is asked to assess the acceptability of the statement based on general knowledge or contextual plausibility. As~\citet{skovgaard2016relevance} include an instruction paragraph on what exactly is meant by acceptability in their task, a summarized version of this is also included in the LLM prompts for acceptability:

\begin{quote}
    If a statement is considered "acceptable", it means that the information expressed by it is reasonable given some context or prior knowledge about the world.
\end{quote}

\paragraph{Few-shot and CoT} \label{app_fewshot_cot}

While the vanilla prompts follow a zero-shot format, providing only the task instruction, the scenario context, and the input item, the few-shot variants expand on this by including three illustrative examples after the instruction. As shown in Figure~\ref{fig_llm_prompt_fewshot_examples}, these examples are embedded within a single, shared context and represent a range of output values (one high, one moderate, and one low) to exemplify the scale of possible responses.

The example context is manually constructed to closely resemble the scenario style used in the original study: it establishes that a character, "Luisa", wants to pick up some groceries after work, and then adds constraints on the situation by stating that the supermarket is small and might not carry all the necessary items. The example inputs (e.g., \emph{If Luisa drives to a bigger supermarket, she will find all the ingredients she needs}) are randomly selected to represent diverse output levels without biasing towards specific relation types. Moreover, relation types are deliberately omitted from the prompt to prevent influencing the model’s judgments.
Those examples, designed to illustrate a clear probability gradient within the example scenario, remain identical across all prompts. Since the example output values are primarily intended as proxies to indicate a realistic range of responses, minor variations are introduced between judgment types (e.g., $92 \to 94 \to 95$ for the high example) to avoid repetition.
Additionally, as shown in Figure~\ref{fig_llm_prompt_cot_examples}, the CoT prompts also include step-by-step reasoning that exemplifies how to arrive at the final judgment score.

\begin{table*}[tbp]
\centering
\small
\renewcommand{\arraystretch}{1.15}
\setlength{\tabcolsep}{4pt}

\begin{tabularx}{\linewidth}{
  @{}
  >{\raggedright\arraybackslash}p{0.18\linewidth}
  >{\raggedright\arraybackslash}p{0.18\linewidth}
  >{\raggedright\arraybackslash}X
  >{\raggedright\arraybackslash}X
  @{}}
\toprule
\textbf{Method} & \textbf{How It Works} & \textbf{Advantages} & \textbf{Disadvantages} \\
\midrule

\textbf{Direct elicitation} (ask the model ``What is $P(B \mid A)$?'')
&
Prompt the model to output a numeric probability for $B$ under the condition $A$ (optionally after reasoning)
&
\begin{minipage}[t]{\linewidth}\vspace{-6pt}
\begin{itemize}[leftmargin=*,topsep=0pt,partopsep=0pt,itemsep=0pt,parsep=0pt]
  \item Uses global semantics \& latent knowledge; can consider $A$ and $B$ holistically rather than token-by-token
  \item Supports reasoning (e.g., via CoT) before committing to a number.
  \item Black-box friendly: no logit access required
  \item Empirically, verbalized probabilities can be well-calibrated—often better than token-likelihoods for instruct/chat models~\citep{tian-etal-2023-just}
\end{itemize}
\end{minipage}
&
\begin{minipage}[t]{\linewidth}\vspace{-6pt}
\begin{itemize}[leftmargin=*,topsep=0pt,partopsep=0pt,itemsep=0pt,parsep=0pt]
  \item Less decomposable/explainable at the score level (gives a number, not a token-level breakdown)
  \item Variance under sampling (for non-greedy decoding)
\end{itemize}
\end{minipage}
\\
\addlinespace[2pt]

\textbf{Sentence probability} (likelihood of text $B$ given text $A$)
&
Compute $P(B \mid A)$ by multiplying (summing log) token probabilities of the exact string $B$ under left-to-right factorization
&
\begin{minipage}[t]{\linewidth}\vspace{-6pt}
\begin{itemize}[leftmargin=*,topsep=0pt,partopsep=0pt,itemsep=0pt,parsep=0pt]
  \item Well-defined and deterministic (given fixed model \& settings)
  \item Decomposable at token level; supports log-likelihood ratios and ablations.
  \item Aligns with the model’s pre-training objective (good for base models)
\end{itemize}
\end{minipage}
&
\begin{minipage}[t]{\linewidth}\vspace{-6pt}
\begin{itemize}[leftmargin=*,topsep=0pt,partopsep=0pt,itemsep=0pt,parsep=0pt]
  \item Measures string form, not pure semantics; highly paraphrase- and tokenization-sensitive
  \item Length bias (longer $B$ penalized); normalization choices matter (per-token, per-byte, etc.)
  \item Not directly comparable across models without normalization (different tokenizers, logit scales, priors)
  \item Requires logit access
  \item Can’t natively leverage the model’s reasoning (e.g., via CoT) during scoring
\end{itemize}
\end{minipage}
\\
\bottomrule
\end{tabularx}
\caption{Comparison of methods for eliciting (conditional) probability judgments with large language models.}
\label{tab_elicitation_methods}
\end{table*}

\subsection{Discussion of Elicitation Methods}\label{app_elicitation_methods}
While we use direct elicitation to retrieve (conditional) probability and acceptability ratings (cf. Sections~\ref{exp_setup} and~\ref{app_models_prompts}), a well-established approach~\citep[e.g.,][]{tian-etal-2023-just, yang2025on}, other methods exist for obtaining a model's probability estimate for a given statement. 
As an alternative to direct elicitation, we also compute the sentence probability a model assigns to a given statement by leveraging its output probability distribution over the model's vocabulary (see Section~\ref{elicitation_methods_main}). 

While both direct elicitation and sentence probability have their respective advantages and disadvantages (as illustrated in Table~\ref{tab_elicitation_methods}), they bear key differences that should be considered. 
Sentence probability quantifies how probable a model considers a given string. This captures surface form rather than semantics and does not necessarily reflect how likely the model deems the statement to be true. Additionally, pure sentence probability is highly sensitive to paraphrasing and tokenization and exhibits a notable length bias. 
In contrast, direct elicitation allows the model to be asked directly for the probability that a given statement is true. For instruct/chat models in particular, direct elicitation has been shown to be better calibrated than token likelihoods~\citep{tian-etal-2023-just}, as it leverages the model’s internal latent knowledge and reasoning.

\subsection{Linear Mixed-Effects Model} \label{app_linear_model}

\paragraph{Linear Mixed-Effects Models}
Put simply, a linear mixed-effects model can be viewed as an extension of linear regression: Whereas simple regression fits a single trend line through the data, linear mixed-effects models estimate an average trend while adjusting for group-level variation (e.g., due to different scenarios). This results in more accurate and generalizable findings, particularly in datasets with multiple co-existing sources of variation~\citep{brown2021introduction, vasishth2010foundations}.

\paragraph{Predictors and Sources of Variation}
The main fixed effects are conditional probability $P(B \mid A)$, relation type (supporting, undermining, unrelated), and judgment metric (acceptability vs. if-probability). To account for potential context-specific variation, random effects are included for scenario and \emph{respondent}, where the latter refers to either a human participant or an LLM sampling iteration.

\begin{table}[b!]
  \small
  \centering
  \begin{tabular}{lccc}
    \toprule
     & \textbf{IRR} & \textbf{NEG} & \textbf{POS} \\
    \midrule
    Llama 8B (vanilla) & \num{-0.367937500} & \num{-0.274062500} & \num{0.006270833} \\
    Llama 70B (vanilla) & \num{-0.2604167} & \num{-0.2663542} & \num{0.0355000} \\
    Qwen 7B (vanilla) & \num{-0.22187500} & \num{-0.23802083} & \num{0.05572917} \\
    Qwen 72B (vanilla) & \num{-0.2718750} & \num{-0.2109375} & \num{0.1603125} \\
    \midrule
    Llama 8B (few-shot) & \num{-0.3286458} & \num{-0.2170208} & \num{0.0663125} \\
    Llama 70B (few-shot) & \num{-0.30275} & \num{-0.22650} & \num{0.11300} \\
    Qwen 7B (few-shot) & \num{-0.19775000} & \num{-0.19418750} & \num{0.09379167} \\
    Qwen 72B (few-shot) & \num{-0.2782708} & \num{-0.2140208} & \num{0.1521458} \\
    \midrule
    Llama 70B (CoT) & \num{-0.2767917} & \num{-0.2275000} & \num{0.1713958} \\
    Qwen 72B (CoT) & \num{-0.3297500} & \num{-0.1986458} & \num{0.1393958} \\
    \bottomrule
  \end{tabular}
  \caption{Means of conditional statement judgments for all LLMs (joint means of both acceptability and if-probability), divided by relation types.}
  \label{tab:general_means}
\end{table}

\paragraph{Model Specification and Adaptation}

Each linear mixed-effects model uses \emph{judgment}, i.e., the judgments of conditional statements, as the dependent variable.
A single model is fit per respondent (LLM or human) with the same structure, enabling later comparisons.
Following best practices~\citep{bates2015fitting, brown2021introduction}, we balance model complexity and convergence. Due to singularity issues with full random-slopes structures, we simplify the models by including random intercepts for both \emph{respondent} (sampling iteration) and scenario in the LLM linear mixed-effects models.  The human mixed-effects model, however, is fit with respondent as a random slope instead, as for humans, this represents the individual participants and is expected to introduce much greater variation.
This choice captures key variance across participants and model samples while retaining model stability.

\subsection{ANOVA Tables} \label{app_anova}
Tables~\ref{tab:anova_Llama 8B (vanilla)} to~\ref{tab:anova_Qwen 72B (CoT)} illustrate the results from the ANOVA analyses.

\begin{table*}
  \centering
  \begin{tabular}{lrrrrr}
    \hline
    \textbf{Effect} & \textbf{F value} & \textbf{Df} & \textbf{Residual Df} & \textbf{$p$-value} \\
    \hline
    (Intercept) & \num{79.1488} & 1 & \num{21.73} & $< 0.0001$ & *** \\
    Conditional probability & \num{43.6288} & 1 & \num{1421.39} & $< 0.0001$ & *** \\
    Relation type & \num{65.6226} & 2 & \num{1414.24} & $< 0.0001$ & *** \\
    Judgment metric (P / A) & \num{4.3749} & 1 & \num{1413.00} & \pnum{0.036651} & * \\
    Cond. prob. $\times$ Relation & \num{7.9795} & 2 & \num{1420.91} & $< 0.001$ & *** \\
    Cond. prob. $\times$ Metric & \num{2.4851} & 1 & \num{1413.00} & \pnum{0.115149} &  \\
    Relation $\times$ Metric & \num{6.0883} & 2 & \num{1413.00} & \pnum{0.002329} & ** \\
    Cond. prob. $\times$ Relation $\times$ Metric & \num{3.7105} & 2 & \num{1413.00} & \pnum{0.024705} & * \\\hline
  \end{tabular}
  \caption{Type III ANOVA table for fixed effects in the mixed-effects model of \emph{Llama 8B (vanilla)} judgments. Probability and acceptability ratings are predicted by conditional probability, relation type, and judgment type. Degrees of freedom are computed using the Kenward--Roger approximation. Significance codes: \textbf{***} $p < .001$, \textbf{**} $p < .01$, \textbf{*} $p < 0.05$.}
  \label{tab:anova_Llama 8B (vanilla)}
\end{table*}

\begin{table*}
  \centering
  \begin{tabular}{lrrrrr}
    \hline
    \textbf{Effect} & \textbf{F value} & \textbf{Df} & \textbf{Residual Df} & \textbf{$p$-value} \\
    \hline
    (Intercept) & \num{35.4628} & 1 & \num{30.41} & $< 0.0001$ & *** \\
    Conditional probability & \num{244.5425} & 1 & \num{1422.38} & $< 0.0001$ & *** \\
    Relation type & \num{2.8160} & 2 & \num{1415.34} & \pnum{0.06018} & . \\
    Judgment metric (P / A) & \num{3.6091} & 1 & \num{1413.00} & \pnum{0.05767} & . \\
    Cond. prob. $\times$ Relation & \num{0.5262} & 2 & \num{1418.53} & \pnum{0.59097} &  \\
    Cond. prob. $\times$ Metric & \num{2.2972} & 1 & \num{1413.00} & \pnum{0.12983} &  \\
    Relation $\times$ Metric & \num{0.2448} & 2 & \num{1413.00} & \pnum{0.78289} &  \\
    Cond. prob. $\times$ Relation $\times$ Metric & \num{0.1502} & 2 & \num{1413.00} & \pnum{0.86055} &  \\\hline
  \end{tabular}
  \caption{Type III ANOVA table for fixed effects in the mixed-effects model of \emph{Llama 70B (vanilla)} judgments. Probability and acceptability ratings are predicted by conditional probability, relation type, and judgment type. Degrees of freedom are computed using the Kenward--Roger approximation. Significance codes: \textbf{***} $p < .001$, \textbf{**} $p < .01$, \textbf{*} $p < 0.05$.}
  \label{tab:anova_Llama 70B (vanilla)}
\end{table*}

\begin{table*}
  \centering
  \begin{tabular}{lrrrrr}
    \hline
    \textbf{Effect} & \textbf{F value} & \textbf{Df} & \textbf{Residual Df} & \textbf{$p$-value} \\
    \hline
    (Intercept) & \num{72.0300} & 1 & \num{16.77} & $< 0.0001$ & *** \\
    Conditional probability & \num{94.3585} & 1 & \num{1416.96} & $< 0.0001$ & *** \\
    Relation type & \num{100.0870} & 2 & \num{1413.35} & $< 0.0001$ & *** \\
    Judgment metric (P / A) & \num{5.9821} & 1 & \num{1413.00} & \pnum{0.01457} & * \\
    Cond. prob. $\times$ Relation & \num{10.7228} & 2 & \num{1417.78} & $< 0.0001$ & *** \\
    Cond. prob. $\times$ Metric & \num{24.0222} & 1 & \num{1413.00} & $< 0.0001$ & *** \\
    Relation $\times$ Metric & \num{1.5142} & 2 & \num{1413.00} & \pnum{0.22033} &  \\
    Cond. prob. $\times$ Relation $\times$ Metric & \num{0.6884} & 2 & \num{1413.00} & \pnum{0.50253} &  \\\hline
  \end{tabular}
  \caption{Type III ANOVA table for fixed effects in the mixed-effects model of \emph{Qwen 7B (vanilla)} judgments. Probability and acceptability ratings are predicted by conditional probability, relation type, and judgment type. Degrees of freedom are computed using the Kenward--Roger approximation. Significance codes: \textbf{***} $p < .001$, \textbf{**} $p < .01$, \textbf{*} $p < 0.05$.}
  \label{tab:anova_Qwen 7B (vanilla)}
\end{table*}

\begin{table*}
  \centering
  \begin{tabular}{lrrrrr}
    \hline
    \textbf{Effect} & \textbf{F value} & \textbf{Df} & \textbf{Residual Df} & \textbf{$p$-value} \\
    \hline
    (Intercept) & \num{224.6492} & 1 & \num{18.65} & $< 0.0001$ & *** \\
    Conditional probability & \num{34.0262} & 1 & \num{1414.81} & $< 0.0001$ & *** \\
    Relation type & \num{200.9049} & 2 & \num{1414.06} & $< 0.0001$ & *** \\
    Judgment metric (P / A) & \num{3.3297} & 1 & \num{1413.00} & \pnum{0.0682523} & . \\
    Cond. prob. $\times$ Relation & \num{59.3747} & 2 & \num{1416.95} & $< 0.0001$ & *** \\
    Cond. prob. $\times$ Metric & \num{72.7878} & 1 & \num{1413.00} & $< 0.0001$ & *** \\
    Relation $\times$ Metric & \num{0.4891} & 2 & \num{1413.00} & \pnum{0.6132904} &  \\
    Cond. prob. $\times$ Relation $\times$ Metric & \num{7.5252} & 2 & \num{1413.00} & \pnum{0.0005612} & *** \\\hline
  \end{tabular}
  \caption{Type III ANOVA table for fixed effects in the mixed-effects model of \emph{Qwen 72B (vanilla)} judgments. Probability and acceptability ratings are predicted by conditional probability, relation type, and judgment type. Degrees of freedom are computed using the Kenward--Roger approximation. Significance codes: \textbf{***} $p < .001$, \textbf{**} $p < .01$, \textbf{*} $p < 0.05$.}
  \label{tab:anova_Qwen 72B (vanilla)}
\end{table*}

\begin{table*}
  \centering
  \begin{tabular}{lrrrrr}
    \hline
    \textbf{Effect} & \textbf{F value} & \textbf{Df} & \textbf{Residual Df} & \textbf{$p$-value} \\
    \hline
    (Intercept) & \num{79.6736} & 1 & \num{27.72} & $< 0.0001$ & *** \\
    Conditional probability & \num{29.6153} & 1 & \num{1423.62} & $< 0.0001$ & *** \\
    Relation type & \num{68.2008} & 2 & \num{1415.28} & $< 0.0001$ & *** \\
    Judgment metric (P / A) & \num{15.8801} & 1 & \num{1413.00} & $< 0.0001$ & *** \\
    Cond. prob. $\times$ Relation & \num{9.8630} & 2 & \num{1420.44} & $< 0.0001$ & *** \\
    Cond. prob. $\times$ Metric & \num{8.1865} & 1 & \num{1413.00} & \pnum{0.0042825} & ** \\
    Relation $\times$ Metric & \num{7.7382} & 2 & \num{1413.00} & $< 0.001$ & *** \\
    Cond. prob. $\times$ Relation $\times$ Metric & \num{1.9840} & 2 & \num{1413.00} & \pnum{0.1379067} &  \\\hline
  \end{tabular}
  \caption{Type III ANOVA table for fixed effects in the mixed-effects model of \emph{Llama 8B (few-shot)} judgments. Probability and acceptability ratings are predicted by conditional probability, relation type, and judgment type. Degrees of freedom are computed using the Kenward--Roger approximation. Significance codes: \textbf{***} $p < .001$, \textbf{**} $p < .01$, \textbf{*} $p < 0.05$.}
  \label{tab:anova_Llama 8B (few-shot)}
\end{table*}

\begin{table*}
  \centering
  \begin{tabular}{lrrrrr}
    \hline
    \textbf{Effect} & \textbf{F value} & \textbf{Df} & \textbf{Residual Df} & \textbf{$p$-value} \\
    \hline
    (Intercept) & \num{83.1165} & 1 & \num{18.76} & $< 0.0001$ & *** \\
    Conditional probability & \num{202.7056} & 1 & \num{1419.33} & $< 0.0001$ & *** \\
    Relation type & \num{63.5739} & 2 & \num{1414.17} & $< 0.0001$ & *** \\
    Judgment metric (P / A) & \num{14.2564} & 1 & \num{1413.00} & $< 0.001$ & *** \\
    Cond. prob. $\times$ Relation & \num{28.0766} & 2 & \num{1416.22} & $< 0.0001$ & *** \\
    Cond. prob. $\times$ Metric & \num{11.9243} & 1 & \num{1413.00} & \pnum{0.0005704} & *** \\
    Relation $\times$ Metric & \num{2.0273} & 2 & \num{1413.00} & \pnum{0.1320761} &  \\
    Cond. prob. $\times$ Relation $\times$ Metric & \num{1.4163} & 2 & \num{1413.00} & \pnum{0.2429547} &  \\\hline
  \end{tabular}
  \caption{Type III ANOVA table for fixed effects in the mixed-effects model of \emph{Llama 70B (few-shot)} judgments. Probability and acceptability ratings are predicted by conditional probability, relation type, and judgment type. Degrees of freedom are computed using the Kenward--Roger approximation. Significance codes: \textbf{***} $p < .001$, \textbf{**} $p < .01$, \textbf{*} $p < 0.05$.}
  \label{tab:anova_Llama 70B (few-shot)}
\end{table*}

\begin{table*}
  \centering
  \begin{tabular}{lrrrrr}
    \hline
    \textbf{Effect} & \textbf{F value} & \textbf{Df} & \textbf{Residual Df} & \textbf{$p$-value} \\
    \hline
    (Intercept) & \num{83.5511} & 1 & \num{22.01} & $< 0.0001$ & *** \\
    Conditional probability & \num{140.3733} & 1 & \num{1422.45} & $< 0.0001$ & *** \\
    Relation type & \num{51.5798} & 2 & \num{1414.02} & $< 0.0001$ & *** \\
    Judgment metric (P / A) & \num{6.0225} & 1 & \num{1413.00} & \pnum{0.01424} & * \\
    Cond. prob. $\times$ Relation & \num{5.8707} & 2 & \num{1419.68} & \pnum{0.00289} & ** \\
    Cond. prob. $\times$ Metric & \num{5.0550} & 1 & \num{1413.00} & \pnum{0.02471} & * \\
    Relation $\times$ Metric & \num{4.2015} & 2 & \num{1413.00} & \pnum{0.01516} & * \\
    Cond. prob. $\times$ Relation $\times$ Metric & \num{3.9840} & 2 & \num{1413.00} & \pnum{0.01882} & * \\\hline
  \end{tabular}
  \caption{Type III ANOVA table for fixed effects in the mixed-effects model of \emph{Qwen 7B (few-shot)} judgments. Probability and acceptability ratings are predicted by conditional probability, relation type, and judgment type. Degrees of freedom are computed using the Kenward--Roger approximation. Significance codes: \textbf{***} $p < .001$, \textbf{**} $p < .01$, \textbf{*} $p < 0.05$.}
  \label{tab:anova_Qwen 7B (few-shot)}
\end{table*}

\begin{table*}
  \centering
  \begin{tabular}{lrrrrr}
    \hline
    \textbf{Effect} & \textbf{F value} & \textbf{Df} & \textbf{Residual Df} & \textbf{$p$-value} \\
    \hline
    (Intercept) & \num{307.3136} & 1 & \num{19.69} & $< 0.0001$ & *** \\
    Conditional probability & \num{72.9970} & 1 & \num{1417.90} & $< 0.0001$ & *** \\
    Relation type & \num{175.0968} & 2 & \num{1414.31} & $< 0.0001$ & *** \\
    Judgment metric (P / A) & \num{42.2393} & 1 & \num{1413.00} & $< 0.0001$ & *** \\
    Cond. prob. $\times$ Relation & \num{48.4982} & 2 & \num{1416.94} & $< 0.0001$ & *** \\
    Cond. prob. $\times$ Metric & \num{59.2902} & 1 & \num{1413.00} & $< 0.0001$ & *** \\
    Relation $\times$ Metric & \num{5.5115} & 2 & \num{1413.00} & \pnum{0.004127} & ** \\
    Cond. prob. $\times$ Relation $\times$ Metric & \num{6.9027} & 2 & \num{1413.00} & \pnum{0.001039} & ** \\\hline
  \end{tabular}
  \caption{Type III ANOVA table for fixed effects in the mixed-effects model of \emph{Qwen 72B (few-shot)} judgments. Probability and acceptability ratings are predicted by conditional probability, relation type, and judgment type. Degrees of freedom are computed using the Kenward--Roger approximation. Significance codes: \textbf{***} $p < .001$, \textbf{**} $p < .01$, \textbf{*} $p < 0.05$.}
  \label{tab:anova_Qwen 72B (few-shot)}
\end{table*}

\begin{table*}
  \centering
  \begin{tabular}{lrrrrr}
    \hline
    \textbf{Effect} & \textbf{F value} & \textbf{Df} & \textbf{Residual Df} & \textbf{$p$-value} \\
    \hline
    (Intercept) & \num{175.7068} & 1 & \num{19.94} & $< 0.0001$ & *** \\
    Conditional probability & \num{165.3692} & 1 & \num{1416.16} & $< 0.0001$ & *** \\
    Relation type & \num{130.9670} & 2 & \num{1414.13} & $< 0.0001$ & *** \\
    Judgment metric (P / A) & \num{23.2551} & 1 & \num{1413.00} & $< 0.0001$ & *** \\
    Cond. prob. $\times$ Relation & \num{25.8703} & 2 & \num{1416.33} & $< 0.0001$ & *** \\
    Cond. prob. $\times$ Metric & \num{18.6004} & 1 & \num{1413.00} & $< 0.0001$ & *** \\
    Relation $\times$ Metric & \num{2.5410} & 2 & \num{1413.00} & \pnum{0.079145} & . \\
    Cond. prob. $\times$ Relation $\times$ Metric & \num{6.1775} & 2 & \num{1413.00} & \pnum{0.002132} & ** \\\hline
  \end{tabular}
  \caption{Type III ANOVA table for fixed effects in the mixed-effects model of \emph{Llama 70B (CoT)} judgments. Probability and acceptability ratings are predicted by conditional probability, relation type, and judgment type. Degrees of freedom are computed using the Kenward--Roger approximation. Significance codes: \textbf{***} $p < .001$, \textbf{**} $p < .01$, \textbf{*} $p < 0.05$.}
  \label{tab:anova_Llama 70B (CoT)}
\end{table*}

\begin{table*}
  \centering
  \begin{tabular}{lrrrrr}
    \hline
    \textbf{Effect} & \textbf{F value} & \textbf{Df} & \textbf{Residual Df} & \textbf{$p$-value} \\
    \hline
    (Intercept) & \num{312.6325} & 1 & \num{20.73} & $< 0.0001$ & *** \\
    Conditional probability & \num{24.4254} & 1 & \num{1416.32} & $< 0.0001$ & *** \\
    Relation type & \num{188.0044} & 2 & \num{1413.85} & $< 0.0001$ & *** \\
    Judgment metric (P / A) & \num{14.4341} & 1 & \num{1413.00} & $< 0.001$ & *** \\
    Cond. prob. $\times$ Relation & \num{65.2923} & 2 & \num{1416.77} & $< 0.0001$ & *** \\
    Cond. prob. $\times$ Metric & \num{21.0905} & 1 & \num{1413.00} & $< 0.0001$ & *** \\
    Relation $\times$ Metric & \num{2.4979} & 2 & \num{1413.00} & \pnum{0.0826179} & . \\
    Cond. prob. $\times$ Relation $\times$ Metric & \num{1.5992} & 2 & \num{1413.00} & \pnum{0.2024169} &  \\\hline
  \end{tabular}
  \caption{Type III ANOVA table for fixed effects in the mixed-effects model of \emph{Qwen 72B (CoT)} judgments. Probability and acceptability ratings are predicted by conditional probability, relation type, and judgment type. Degrees of freedom are computed using the Kenward--Roger approximation. Significance codes: \textbf{***} $p < .001$, \textbf{**} $p < .01$, \textbf{*} $p < 0.05$.}
  \label{tab:anova_Qwen 72B (CoT)}
\end{table*}

\begin{table*}
  \centering
  \begin{tabular}{lrrrrr}
    \hline
    \textbf{Effect} & \textbf{F value} & \textbf{Df} & \textbf{Residual Df} & \textbf{$p$-value} \\
    \hline
    (Intercept) & \num{190.6202} & 1 & \num{213.73} & $< 0.0001$ & *** \\
    Conditional probability & \num{80.7620} & 1 & \num{234.87} & $< 0.0001$ & *** \\
    Relation type & \num{116.1519} & 2 & \num{235.95} & $< 0.0001$ & *** \\
    Judgment metric (P / A) & \num{1.9410} & 1 & \num{1190.22} & \pnum{0.1638} &  \\
    Cond. prob. $\times$ Relation & \num{22.7321} & 2 & \num{238.67} & $< 0.0001$ & *** \\
    Cond. prob. $\times$ Metric & \num{1.1194} & 1 & \num{655.91} & \pnum{0.2904} &  \\
    Relation $\times$ Metric & \num{2.0725} & 2 & \num{786.15} & \pnum{0.1266} &  \\
    Cond. prob. $\times$ Relation $\times$ Metric & \num{1.0946} & 2 & \num{641.12} & \pnum{0.3353} &  \\\hline
  \end{tabular}
  \caption{Type III ANOVA table for fixed effects in the mixed-effects model of \emph{Human} judgments. Probability and acceptability ratings are predicted by conditional probability, relation type, and judgment type. Degrees of freedom are computed using the Kenward--Roger approximation. Significance codes: \textbf{***} $p < .001$, \textbf{**} $p < .01$, \textbf{*} $p < 0.05$.}
  \label{tab:anova_Human}
\end{table*}

\begin{table*}
  \small
  \centering
  \begin{tabular}{lrrrrrr}
    \hline
     & & & & & &  \\
    \textbf{Low Cond. Prob.} & \textbf{IRR $\times$ NEG} & & \textbf{IRR $\times$ POS} & & \textbf{NEG $\times$ POS} & \\
    \hline
     & \textbf{Estimate} & \textbf{$p$} & \textbf{Estimate} & \textbf{$p$} & \textbf{Estimate} & \textbf{$p$}\\
    \hline
    Llama 8B (vanilla) & \num{-0.0871} & $< 0.001$ & \num{-0.1294} & $< 0.001$ & \num{-0.0423} & \pnum{0.1480} \\
    Llama 70B (vanilla) & \num{-0.07379} & \pnum{0.0090} & \num{0.00701} & \pnum{0.8271} & \num{0.08080} & \pnum{0.0654} \\
    Qwen 7B (vanilla) & \num{-0.00861} & \pnum{0.6840} & \num{-0.13079} & $< 0.0001$ & \num{-0.12218} & $< 0.0001$ \\
    Qwen 72B (vanilla) & \num{-0.0229} & \pnum{0.2323} & \num{-0.1271} & $< 0.0001$ & \num{-0.1042} & $< 0.0001$ \\
    \hline
    Llama 8B (few-shot)& \num{-0.0570} & \pnum{0.0299} & \num{-0.1627} & $< 0.0001$ & \num{-0.1057} & \pnum{0.0009} \\
    Llama 70B (few-shot)& \num{-0.07132} & \pnum{0.0007} & \num{-0.07484} & \pnum{0.0245} & \num{-0.00352} & \pnum{0.9058} \\
    Qwen 7B (few-shot)& \num{-0.05305} & \pnum{0.0631} & \num{-0.17916} & $< 0.0001$ & \num{-0.12611} & $< 0.0001$ \\
    Qwen 72B (few-shot)& \num{-0.02716} & \pnum{0.2629} & \num{-0.10526} & $< 0.001$ & \num{-0.07810} & \pnum{0.0067} \\
    \hline
    Llama 70B (CoT)& \num{-0.1143} & $< 0.0001$ & \num{-0.2352} & $< 0.0001$ & \num{-0.1209} & $< 0.001$ \\
    Qwen 72B (CoT)& \num{-0.0123} & \pnum{1.0000} & \num{-0.1010} & \pnum{0.0025} & \num{-0.0887} & \pnum{0.0030} \\
    \hline
    Human & \num{0.0123} & \pnum{0.4214} & \num{-0.1332} & $< 0.0001$ & \num{-0.1455} & $< 0.0001$ \\\hline
    \hline
     & & & & & & \\
    \textbf{Medium C. Prob.} & \textbf{IRR $\times$ NEG} & & \textbf{IRR $\times$ POS} & & \textbf{NEG $\times$ POS} & \\
    \hline
     & \textbf{Estimate} & \textbf{$p$} & \textbf{Estimate} & \textbf{$p$} & \textbf{Estimate} & \textbf{$p$}\\
    \hline
    Llama 8B (vanilla) & \num{-0.1374} & $< 0.0001$ & \num{-0.2679} & $< 0.0001$ & \num{-0.1305} & $< 0.0001$ \\
    Llama 70B (vanilla) & \num{-0.08605} & \pnum{0.0009} & \num{-0.02162} & \pnum{0.5978} & \num{0.06442} & \pnum{0.0328} \\
    Qwen 7B (vanilla) & \num{-0.07427} & $< 0.0001$ & \num{-0.24187} & $< 0.0001$ & \num{-0.16760} & $< 0.0001$ \\
    Qwen 72B (vanilla) & \num{-0.2110} & $< 0.0001$ & \num{-0.3005} & $< 0.0001$ & \num{-0.0895} & $< 0.0001$ \\
    \hline
    Llama 8B (few-shot)& \num{-0.1639} & $< 0.0001$ & \num{-0.2550} & $< 0.0001$ & \num{-0.0910} & $< 0.0001$ \\
    Llama 70B (few-shot)& \num{-0.25413} & $< 0.0001$ & \num{-0.16013} & $< 0.0001$ & \num{0.09400} & $< 0.0001$ \\
    Qwen 7B (few-shot)& \num{-0.17635} & $< 0.0001$ & \num{-0.23884} & $< 0.0001$ & \num{-0.06249} & \pnum{0.0012} \\
    Qwen 72B (few-shot)& \num{-0.20230} & $< 0.0001$ & \num{-0.24533} & $< 0.0001$ & \num{-0.04303} & \pnum{0.0038} \\
    \hline
    Llama 70B (CoT)& \num{-0.2692} & $< 0.0001$ & \num{-0.2825} & $< 0.0001$ & \num{-0.0132} & \pnum{0.4593} \\
    Qwen 72B (CoT)& \num{-0.2683} & $< 0.0001$ & \num{-0.3193} & $< 0.0001$ & \num{-0.0510} & \pnum{0.0022} \\
    \hline
    Human & \num{-0.0809} & $< 0.0001$ & \num{-0.3242} & $< 0.0001$ & \num{-0.2433} & $< 0.0001$ \\\hline
    \hline
     & & & & & & \\
    \textbf{High Cond. Prob.} & \textbf{IRR $\times$ NEG} & & \textbf{IRR $\times$ POS} & & \textbf{NEG $\times$ POS} & \\
    \hline
     & \textbf{Estimate} & \textbf{$p$} & \textbf{Estimate} & \textbf{$p$} & \textbf{Estimate} & \textbf{$p$}\\
    \hline
    Llama 8B (vanilla) & \num{-0.1878} & $< 0.0001$ & \num{-0.4064} & $< 0.0001$ & \num{-0.2186} & $< 0.0001$ \\
    Llama 70B (vanilla) & \num{-0.09831} & \pnum{0.0654} & \num{-0.05026} & \pnum{0.3313} & \num{0.04805} & \pnum{0.5395} \\
    Qwen 7B (vanilla) & \num{-0.13993} & $< 0.0001$ & \num{-0.35294} & $< 0.0001$ & \num{-0.21302} & $< 0.0001$ \\
    Qwen 72B (vanilla) & \num{-0.3990} & $< 0.0001$ & \num{-0.4738} & $< 0.0001$ & \num{-0.0747} & \pnum{0.0005} \\
    \hline
    Llama 8B (few-shot)& \num{-0.2708} & $< 0.0001$ & \num{-0.3472} & $< 0.0001$ & \num{-0.0763} & \pnum{0.0364} \\
    Llama 70B (few-shot)& \num{-0.43693} & $< 0.0001$ & \num{-0.24542} & $< 0.0001$ & \num{0.19151} & $< 0.0001$ \\
    Qwen 7B (few-shot)& \num{-0.29965} & $< 0.0001$ & \num{-0.29852} & $< 0.0001$ & \num{0.00112} & \pnum{0.9724} \\
    Qwen 72B (few-shot)& \num{-0.37744} & $< 0.0001$ & \num{-0.38540} & $< 0.0001$ & \num{-0.00796} & \pnum{0.7142} \\
    \hline
    Llama 70B (CoT)& \num{-0.4242} & $< 0.0001$ & \num{-0.3297} & $< 0.0001$ & \num{0.0945} & \pnum{0.0010} \\
    Qwen 72B (CoT)& \num{-0.5242} & $< 0.0001$ & \num{-0.5375} & $< 0.0001$ & \num{-0.0133} & \pnum{1.0000} \\
    \hline
    Human & \num{-0.1741} & $< 0.0001$ & \num{-0.5152} & $< 0.0001$ & \num{-0.3410} & $< 0.0001$ \\\hline
  \end{tabular}
  \caption{Pairwise comparisons of estimated marginal means at low, medium, and high conditional probability for all LLMs and humans. SE: $0.01 - 0.04$ for all. Statistical significance at $p < 0.05$.}
  \label{tab:emm}
\end{table*}

\begin{figure*}[ht]
 \centering
  \includegraphics[width=0.9\linewidth]{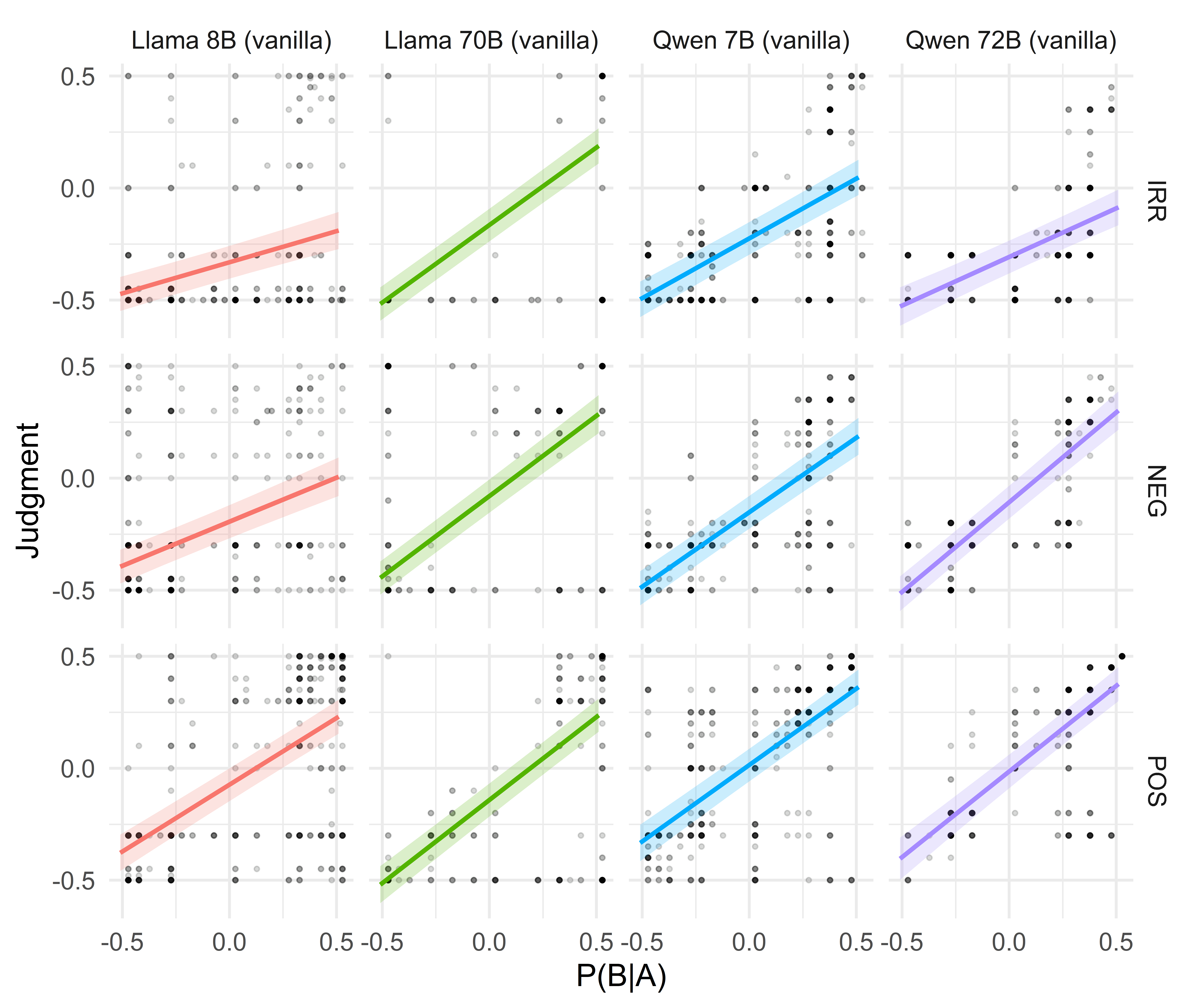}
  \caption{Scatterplots of data points and trend lines for all vanilla LLMs, divided by relation type.}
  \label{plot:scatter_trendlines_vanilla}
\end{figure*}

\newpage

\begin{figure*}[t!]
  \centering
  \begin{tcolorbox}[
    colback=MidnightBlue!5!white,
    colframe=MidnightBlue,
    boxrule=0.4pt,
    rounded corners,
    width=\textwidth,
    before upper={\parindent0pt},
    fontupper=\small
  ]
\#\#\# Instruction:

\vspace{1em}

You will be given an assumption and a sentence: \\
- Assumption: a supposition to treat as true \\
- Sentence: the claim to evaluate

\vspace{1em}

Given the assumption, how probable is it that the sentence is true? Your answer must be a single number between 0 and 100, where: \\
- 0 means "completely improbable" \\
- 100 means "highly probable"

\vspace{1em}

Respond only with this number and nothing else.

\vspace{1em}

\#\#\# Output Format:

\vspace{1em}

Answer: <number>

\vspace{1em}

\#\#\# Input: \\
Context: Nicole’s skin is pale and she wants a suntan with a mild brown color. It is summer, the weather is nice, and there is a beach nearby. Unfortunately, Nicole’s skin is very sensitive to the sun, and she is somewhat bad at controlling the amount of her exposure to the sun.

\vspace{1em}

Assumption: Suppose Nicole sunbathes on the beach. \\
Sentence: She will get sunburned
  \end{tcolorbox}
  \caption{Prompt for eliciting conditional probability judgments (vanilla version).}
  \label{fig_llm_prompt_vanilla_cprob}
\end{figure*}

\begin{figure*}[ht]
  \centering
  \begin{tcolorbox}[
    colback=MidnightBlue!5!white,
    colframe=MidnightBlue,
    boxrule=0.4pt,
    rounded corners,
    width=\textwidth,
    before upper={\parindent0pt},
    fontupper=\small
  ]
\#\#\# Instruction:

\vspace{1em}

Rate the probability of the following statement. Your answer must be a single number between 0 and 100, where: \\
- 0 means "completely improbable" \\
- 100 means "highly probable"

\vspace{1em}

Respond only with this number and nothing else.

\vspace{1em}

\#\#\# Output Format:

\vspace{1em}

Answer: <number>

\vspace{1em}

\#\#\# Input: \\
Context: Nicole’s skin is pale and she wants a suntan with a mild brown color. It is summer, the weather is nice, and there is a beach nearby. Unfortunately, Nicole’s skin is very sensitive to the sun, and she is somewhat bad at controlling the amount of her exposure to the sun.

\vspace{1em}

Statement: If Nicole sunbathes on the beach, then she will get sunburned.
  \end{tcolorbox}
  \caption{Prompt for eliciting If $A$, then $B$ probability judgments (vanilla version).}
  \label{fig_llm_prompt_vanilla_ifprob}
\end{figure*}

\begin{figure*}[ht]
  \centering
  \begin{tcolorbox}[
    colback=MidnightBlue!5!white,
    colframe=MidnightBlue,
    boxrule=0.4pt,
    rounded corners,
    width=\textwidth,
    before upper={\parindent0pt},
    fontupper=\small
  ]
\#\#\# Instruction:

If a statement is considered "acceptable", it means that the information expressed by it is reasonable given some context or prior knowledge about the world.

\vspace{1em}

Rate the acceptability of the following statement. Your answer must be a single number between 0 and 100, where: \\
- 0 means "completely unacceptable" \\
- 100 means "highly acceptable"

\vspace{1em}

Respond only with this number and nothing else.

\vspace{1em}

\#\#\# Output Format:

\vspace{1em}

Answer: <number>

\vspace{1em}

\#\#\# Input: \\
Context: Nicole’s skin is pale and she wants a suntan with a mild brown color. It is summer, the weather is nice, and there is a beach nearby. Unfortunately, Nicole’s skin is very sensitive to the sun, and she is somewhat bad at controlling the amount of her exposure to the sun.

\vspace{1em}

Statement: If Nicole sunbathes on the beach, then she will get sunburned.
  \end{tcolorbox}
  \caption{Prompt for eliciting If $A$, then $B$ acceptability judgments (vanilla version).}
  \label{fig_llm_prompt_vanilla_ifacc}
\end{figure*}

\begin{figure*}[ht]
  \centering
  \begin{tcolorbox}[
    colback=MidnightBlue!5!white,
    colframe=MidnightBlue,
    boxrule=0.4pt,
    rounded corners,
    width=\textwidth,
    before upper={\parindent0pt},
    fontupper=\small
  ]
Context: Luisa wants to pick up some groceries on her way home from work because she needs the ingredients to cook her favorite meal tonight. However, the supermarket is very small and does not always carry all the things that other supermarkets do. Driving to a bigger supermarket would take a very long time, but she is already pretty hungry.

\vspace{1em}

Example 1 (High probability):

\vspace{1em}

Statement: If Luisa drives to a bigger supermarket, she will find all the ingredients she needs. \\
Answer: 92

\vspace{1em}

Example 2 (Moderate probability):

\vspace{1em}

Statement: If Luisa drives to the small supermarket, she will find all the ingredients she needs. \\
Answer: 57

\vspace{1em}

Example 3 (Low probability):

\vspace{1em}

Statement: If Luisa goes straight home, she will find all the ingredients she needs. \\
Answer: 12
  \end{tcolorbox}
  \caption{Example context and statements used in the few-shot prompt variants.}
  \label{fig_llm_prompt_fewshot_examples}
\end{figure*}

\begin{figure*}[ht]
  \centering
  \begin{tcolorbox}[
    colback=MidnightBlue!5!white,
    colframe=MidnightBlue,
    boxrule=0.4pt,
    rounded corners,
    width=\textwidth,
    before upper={\parindent0pt},
    fontupper=\small
  ]
Example 1 (High probability):

\vspace{1em}

Statement: If Luisa drives to a bigger supermarket, she will find all the ingredients she needs. \\
Reasoning: \\
1. The context tells us that Luisa is looking for specific ingredients to cook her favorite meal. \\
2. Bigger supermarkets are generally well-stocked and carry a wider variety of products. \\
3. It is reasonable to expect that she will find the necessary items there. \\
4. Although there's a small chance that something might be out of stock, this doesn't undermine the general plausibility. \\
5. Therefore, the statement expresses a reasonable and plausible scenario. \\
Answer: 95

\vspace{1em}

Example 2 (Moderate probability):

\vspace{1em}

Statement: If Luisa drives to the small supermarket, she will find all the ingredients she needs. \\
Reasoning: \\
1. The context tells us that the small supermarket does not always carry everything other supermarkets do. \\
2. This implies that its inventory is limited or inconsistent. \\
3. This makes it uncertain whether it has the full range of ingredients she needs. \\
4. While the statement is not impossible, it’s not strongly supported by what we know. \\
5. Therefore, the statement is somewhat acceptable, but questionable. \\
Answer: 51

\vspace{1em}

Example 3 (Low probability):

\vspace{1em}

Statement: If Luisa goes straight home, she will find all the ingredients she needs. \\
Reasoning: \\
1. The context tells us that Luisa wants to stop for groceries because she needs ingredients, implying that she does not already have the ingredients at home. \\
2. If she skips the store and goes straight home, she is unlikely to have everything required for the meal. \\
3. The statement contradicts her reason for going to the store in the first place. \\
4. While there’s always a slim chance she forgot she already has everything, this is implausible. \\
5. Therefore, the statement is largely unacceptable. \\
Answer: 8
  \end{tcolorbox}
  \caption{Example statements and reasoning used in the acceptability CoT prompt variants.}
  \label{fig_llm_prompt_cot_examples}
\end{figure*}

\end{document}